\documentclass{article}

\usepackage[final]{neurips_2025}

\usepackage[utf8]{inputenc} % allow utf-8 input
\usepackage[T1]{fontenc}    % use 8-bit T1 fonts
\usepackage{hyperref}       % hyperlinks
\usepackage{url}            % simple URL typesetting
\usepackage{booktabs}       % professional-quality tables
\usepackage{amsfonts}       % blackboard math symbols
\usepackage{nicefrac}       % compact symbols for 1/2, etc.
\usepackage{microtype}      % microtypography
\usepackage{xcolor}         % colors
\usepackage{amsmath}
\usepackage{amssymb}
\usepackage{mathtools}
\usepackage{amsthm}
\usepackage{tabularx}
\usepackage{caption}
\usepackage{interval}
\usepackage{multirow}
\usepackage{textcmds}
\usepackage{graphicx}
\usepackage{epsfig}

\usepackage{lipsum}

\usepackage{wrapfig}
\usepackage[normalem]{ulem}
\usepackage{algorithm}
\usepackage{algpseudocode}
\usepackage{tikz}
\usepackage{stmaryrd}
\usepackage{xparse}
\usepackage{bm}

\usepackage[autostyle=false, style=english]{csquotes}
\theoremstyle{definition}

\theoremstyle{remark}

\usepackage{xspace}

\usepackage{colortbl}
\usepackage{xcolor}
\usepackage{booktabs}
\usepackage{multirow}
\definecolor{best}{rgb}{1.0, 0.85, 0.85}      % light red
\definecolor{second}{rgb}{1.0, 0.93, 0.75}    % light orange/yellow

\title{UMAMI: Unifying Masked Autoregressive Models and Deterministic Rendering for View Synthesis
}

\newcommand{\mname}{\texttt{UMAMI}\xspace}

\makeatletter
\newcommand{\noThanksMark}{\renewcommand{\@makefnmark}{}}
\makeatother

\author{Thanh-Tung Le$^{*1}$~~
    Tuan Pham$^{*1}$~~
    Tung Nguyen$^{2}$ ~~ \\
    {\bf Deying Kong$^{3}$~~}
    {\bf Xiaohui Xie$^{1\dagger}$~~~} 
    {\bf Stephan Mandt$^{1\dagger}$~~~} \\
    $^1$UCI ~~
    $^2$UCLA ~~
    $^3$Google ~~
}

\begin{document}

\maketitle
\noThanksMark\footnotetext{$^{*}$ These authors contributed equally to this work. $\dagger$ denotes corresponding authors.}

\begin{figure}[h]
\centering
\vspace{-5mm}
\includegraphics[width=\textwidth]{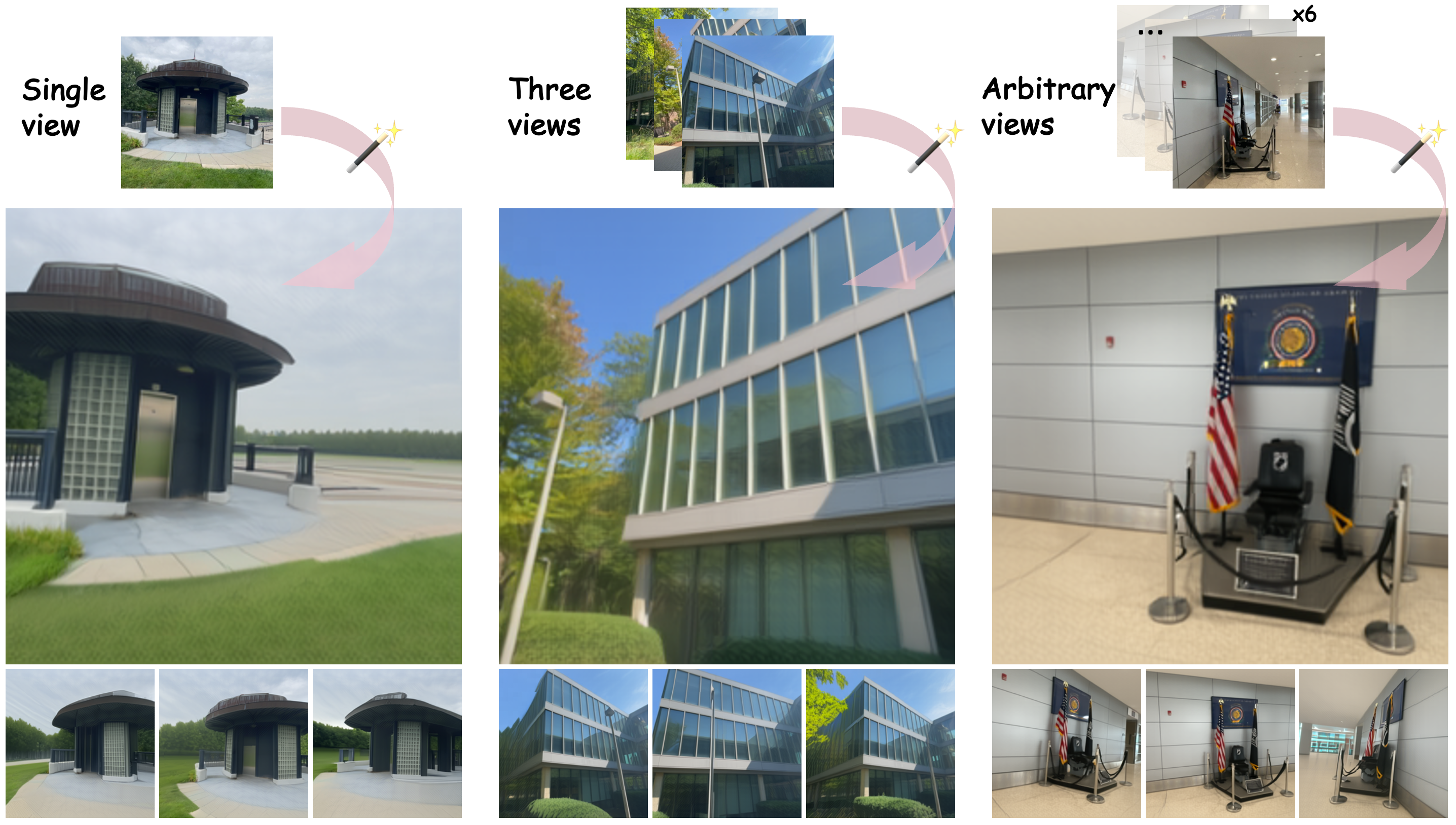}
\caption{\mname synthesizes photorealistic novel views from sparse inputs. Shown are single-view generation, three-view extrapolation, and six-view reconstruction. The hybrid model fuses deterministic rendering with diffusion-based completion for unseen regions, yielding fast and consistent results without explicit 3D priors.}
\label{fig:demo}
\vspace{-3mm}
\end{figure}

\begin{abstract}
  Novel view synthesis (NVS) seeks to render photorealistic, 3D‑consistent images of a scene from unseen camera poses given only a sparse set of posed views. Existing deterministic networks render observed regions quickly but blur unobserved areas, whereas stochastic diffusion‑based methods hallucinate plausible content yet incur heavy training‑ and inference‑time costs. In this paper, we propose a hybrid framework that unifies the strengths of both paradigms. A bidirectional transformer encodes multi‑view image tokens and Plücker‑ray embeddings, producing a shared latent representation. Two lightweight heads then act on this representation: (i) a feed‑forward regression head that renders pixels where geometry is well constrained, and (ii) a masked autoregressive diffusion head that completes occluded or unseen regions. The entire model is trained end‑to‑end with joint photometric and diffusion losses, without handcrafted 3D inductive biases, enabling scalability across diverse scenes. Experiments demonstrate that our method attains state‑of‑the‑art image quality while reducing rendering time by an order of magnitude compared with fully generative baselines.

\end{abstract}

\section{Introduction}
Novel view synthesis (NVS) has long been a central problem in computer vision and graphics, aiming to generate realistic, 3D-consistent images of a scene from new camera viewpoints, using a given set of input views with known poses. Traditional methods often require dense input views, treating NVS as a sequential 3D reconstruction and rendering task~\cite{mildenhall2020nerf, kerbl3Dgaussians}. Recently, modern deep network priors~\cite{yu2020pixelnerf, chen2024mvsplat, gao2024cat3d} have been proposed to address the sparse views reconstruction problems, and achieve realistic rendering results.

Two dominant strategies have emerged for sparse-view NVS using deep networks: deterministic and generative-based methods. Deterministic methods often build generalizable networks that predict novel views by incorporating explicit 3D inductive biases~\cite{yu2020pixelnerf, chen2024mvsplat, chen2021mvsnerf, wang2021ibrnet} or by leveraging priors from large-scale reconstruction models~\cite{hong2024lrmlargereconstructionmodel, li2023instant3d, jin2024lvsmlargeviewsynthesis} with minimum inductive bias. While these approaches can be effective and fast in rendering observed regions, they often struggle with uncertainty in unobserved areas, leading to blurry predictions. Conversely, generative NVS approaches~\cite{4dim, zhou2025stable, gao2024cat3d} can generate plausible content for unseen regions. These methods typically employ pretrained diffusion models conditioned on input views and camera poses. However, despite their strong generative capabilities, they often require extensive training data and computational resources, and their iterative sampling process leads to slow rendering speeds.

In this work, we address the question: \textit{``Can we combine the rendering efficiency of deterministic models with the generation capabilities of generative models?''} We aim to unify these disparate approaches into a single, efficient framework. We observe that conventional diffusion models~\cite{zhou2025stable, gao2024cat3d}, which iteratively generate full images using large UNet or Transformer backbones, can be inefficient if significant portions of the target view are already observable and could be rendered directly by a feed-forward network.

To this end, we introduce \mname, \textbf{U}nifying \textbf{M}asked \textbf{A}utoregressive \textbf{M}odels and Determin\textbf{I}stic Rendering for View Synthesis,  a novel hybrid framework for NVS from sparse inputs. Our approach integrates a masked autoregressive model, trained with a diffusion loss~\cite{li2024autoregressive}, alongside a deterministic rendering head. Specifically, drawing inspiration from recent feed-forward NVS models~\cite{jin2024lvsmlargeviewsynthesis}, we employ a transformer with bidirectional attention. The model encodes input multi-view images tokens and masked target image tokens, conditioned on Plücker ray embeddings for both input and target views to a represenation. The learned representation fulfills a dual role: (1) it conditions a lightweight MLP diffusion backbone that reconstructs unobserved regions through a diffusion loss \cite{li2024autoregressive}, and (2) it serves as input to another MLP that directly renders pixel intensities for observed regions, trained with a photometric loss. Our method is designed to be purely data-driven, minimizing reliance on predefined inductive biases in its representation and rendering. This ``inductive bias-free'' design promotes scalability and generalizability, advantages empirically supported by prior work~\cite{gao2024cat3d, zhou2025stable, jin2024lvsmlargeviewsynthesis}. Ultimately, \mname aims to achieve accurate, training-efficient, and scalable novel view synthesis with photorealistic quality, enjoying both rapid rendering and robust generative completion.

We comprehensively evaluate our model through extensive experiments on RealEstate10K~\cite{zhou2018stereomagnification} and DL3DV \cite{ling2024dl3dv}, demonstrating competitive performance across both interpolation and extrapolation settings, and under varying input-view configurations.

Our contributions as as follows:
\begin{itemize}
    \item A hybrid framework for NVS: We propose \mname, a novel hybrid architecture that combines deterministic and diffusion-based generation to effectively synthesize both visible and occluded regions from sparse views.
    \item We demonstrate that \mname achieves state-of-the-art performance across multiple benchmarks and input settings, while offering favorable trade-offs between speed and quality.
% \vspace{-10mm}
\end{itemize}
\section{Related works}
Novel view synthesis (NVS) is a rapidly advancing field. This section summarizes key prior works most relevant to our approach, with a more exhaustive review provided in the Appendix.
\subsection{Novel view synthesis (NVS)}
Novel view synthesis (NVS) has traditionally relied on image‑based rendering that blends reference views with proxy geometry~\cite{debevec2023modeling, heigl1999plenoptic,sinha2009piecewise}, light‑field techniques that sample the plenoptic function from dense inputs~\cite{davis2012unstructured}, and learning‑based variants that predict blending weights or depth maps with CNNs~\cite{choi2019extreme}. While multiview‑stereo reconstructions enlarge the valid viewing volume~\cite{jancosek2011multi,chaurasia2013depth,penner2017soft}, the breakthrough NeRF model introduced a differentiable volumetric representation whose photometric training signal became the new benchmark for NVS~\cite{mildenhall2020nerf}. Subsequent work has pushed NeRF toward higher fidelity~\cite{barron2021mip,verbin2022refnerf,barron2023zip}, faster inference~\cite{reiser2021kilonerf,hedman2021baking,reiser2023merf}, and fewer input views~\cite{niemeyer2022regnerf,wang2021nerfmm}, or has hybridized it with explicit structures such as dense or sparse voxels~\cite{sun2022dvgo,liu2020nsvf,fridovich2022plenoxels}, low‑rank decompositions and hashing~\cite{chan2022efficient,chen2022tensorf,chen2023factor}, or point/gaussian primitives~\cite{xu2022pointnerf,zhang2022differentiablepoint,feng2022neural,kerbl3Dgaussians}. Despite significant progress in rendering quality, these per-scene optimization methods often suffer from slow training times and limited generalization to novel scenes. 

\subsection{Deterministic NVS}
To address the limitations of per-scene optimization, deterministic NVS methods train a single network across multiple scenes for fast, feed-forward inference. Some approaches, such as PixelSplat~\cite{charatan2024pixelsplat}, MVSplat~\cite{chen2024mvsplat}, and NoPoSplat~\cite{ye2024no}, learn to predict 3D Gaussian parameters directly. While efficient, their reliance on specific 3D representations (e.g., NeRF~\cite{mildenhall2020nerf}, 3D Gaussians~\cite{kerbl3Dgaussians}) can hinder scalability. Alternatively, data-driven methods like LVSM~\cite{jin2024lvsmlargeviewsynthesis} and SRT~\cite{sajjadi2022srt} leverage Transformer-only backbones to map input images and target poses to novel views, demonstrating the potential to synthesize views without explicit 3D representations given sufficient data and careful network design. Although scalable and fast, the deterministic nature of these methods typically restricts view generation to regions observed in the input context. Our method, in contrast, aims to synthesize novel views even when parts of the scene are occluded or outside the context views.

\subsection{Generative NVS}
In addition to deterministic approaches, generative approaches have adapted powerful image and video diffusion models (DMs)~\cite{blattmann2023stable, sd3} for NVS~\cite{zhou2025stable, voleti2024sv3d, 4dim, kong2024eschernet}, leveraging their strong generative priors. Early diffusion-based NVS models~\cite{shi2023mvdream, liu2023syncdreamer, mercier2024hexagen3d} often utilized image DMs conditioned on input images. Contemporary methods increasingly adopt video DMs~\cite{gao2024cat3d, zhou2025stable}, conditioned on camera poses, to achieve finer-grained control and generate high-quality views of unseen regions. However, training these large-scale generative models demands substantial data and computational resources, potentially impacting rendering performance. 
% Our approach uniquely combines the efficiency of deterministic methods for observed regions with the generative capabilities of diffusion models for unobserved areas, achieving both high rendering speed and low computational cost.

\section{Background}
\label{sec:background}

\subsection{Novel View Synthesis}
\label{subsec:pb-NVS}
\textbf{Deterministic approaches} focus on learning a mapping $f_\theta (\textbf{I}^{\text{ctx}}, \bm{\pi}^{\text{ctx}}, \bm{\pi}^{\text{tgt}} )$ that directly generates the target image $\textbf{I}^{\text{tgt}}$. Here, $\textbf{I}^{\text{ctx}}$ and $\textbf{I}^{\text{tgt}}$ represent context and target images, while $\bm{\pi}^{\text{ctx}}$ and $\bm{\pi}^{\text{tgt}}$ denote their respective camera poses. This mapping $f_\theta$ may be realized through pure neural networks~\cite{jin2024lvsmlargeviewsynthesis, suhail2022light} or by integrating 3D inductive biases~\cite{yu2020pixelnerf, charatan2024pixelsplat, chen2024mvsplat, ye2024no}. Although generally efficient, a fundamental limitation of deterministic methods is the inability to generate unseen region due to the deterministic nature.

\textbf{Generative approaches} learn to sample $\textbf{I}^{\text{tgt}}$ from a learned conditional distribution $p_\theta (\textbf{I}^{\text{tgt}} | \textbf{I}^{\text{ctx}}, \bm{\pi}^{\text{ctx}}, \bm{\pi}^{\text{tgt}} )$. This distribution is often modeled using powerful generative frameworks such as diffusion models~\cite{zhou2025stable, gao2024cat3d}. The advantage of such generative techniques lies in their ability to convincingly hallucinate regions absent in the input views. Nevertheless, this capability comes at a significant computational cost for both training and inference, thereby posing challenges to their widespread practical use in NVS applications. 

\subsection{Masked Autoregressive Image Generatation}
\label{subsec:MAR}
Unlike diffusion models, autoregressive (AR) models~\cite{van2016conditional, van2016pixel, gregor2014deep} approach the generation of an ordered token sequence $\{x^1, x^2, \dots, x^n\}$ (with $1 \leq i \leq n$ defining the order) by formulating the problem as ``next token prediction.'' This is mathematically expressed by factorizing the joint probability:
\begin{equation}
    p(x^1, \dots, x^n) = \prod_{i=1}^n p(x^i|x^1, \dots, x^{i-1}),
\end{equation}
where the conditional probability $p(x^i|x^1, \dots, x^{i-1})$ is modeled by a neural network.

Departing from traditional AR methods~\cite{gregor2014deep, van2016pixel}, the Masked Autoregressive (MAR) model~\cite{li2024autoregressive} presents an different approach that unifies random-order AR principles with masked generative modeling through the use of a Diffusion Loss. In MAR, an autoregressive network produces a feature vector $z = f(\cdot) \in \mathbb{R}^D$. This vector, alongside a small MLP $\epsilon_\theta(\cdot)$, is used to model the conditional distribution $p(x|z)$ for a token $x \in \mathbb{R}^d$. The model is trained using the denoising criterion:
\begin{equation}
    \mathcal{L}(z,x) = \mathbb{E}_{\epsilon, t} \left[ ||\epsilon - \epsilon_\theta(x_t | t, z) ||^2 \right],
\end{equation}
where $\epsilon \in \mathbb{R}^d$ is Gaussian noise and $t \in \mathbb{R}$ is the timestep.

Compared to traditional diffusion models~\cite{rombach2022high, peebles2023scalable}, MAR sample an image by iteratively unmasking tokens using the MLP diffusion conditioned on learned latent from transformer. MAR demonstrates computational efficiency while showcasing competitive performance with its counterparts.

Building upon MAR's efficient generative capabilities, our work introduces a novel hybrid method for the NVS task. Specifically, we leverage MAR's generative framework within a hybrid network that incorporates deterministic rendering. Furthermore, we propose a unique sampler specifically designed to efficiently generate novel views, thereby avoiding the iterative full-image generation typical of large backbone architectures~\cite{zhou2025stable, gao2024cat3d}. This approach enables our generative solution to achieve rendering speeds an order of magnitude faster than previous generative NVS methods. We believe this to be the first proposal of a hybrid method that successfully unifies a generative model with a deterministic head to tackle the NVS challenge.
\vspace{-3mm}
\section{Methods}
\label{sec:method}
% For mathematical notations, please refer to our Appendix.
In this section, we first outline our problem formulation (Section~\ref{subsec:prob-form}) and then details our hybrid model (Section~\ref{subsec:hybrid-model}). Subsequently, we present the training loss (Section~\ref{subsec:training-loss}) and conclude by proposing a novel hybrid sampler (Section~\ref{subsec:hybrid-sampler}).

\subsection{Problem Formulation}
\label{subsec:prob-form}
Given sparse input images with known camera poses $\{( \textbf{I}^{\text{ctx}}, \bm{\pi}^{\text{ctx}})\}$, our goal is to model the conditional distribution $p (\textbf{I}^{\text{tgt}} | \textbf{I}^{\text{ctx}}, \bm{\pi}^{\text{ctx}}, \bm{\pi}^{\text{tgt}} )$ to synthesize realistic novel view $\textbf{I}^{\text{tgt}}$ given its camera poses $\bm{\pi}^{\text{tgt}}$. 

\paragraph{Data Representation} To jointly encode image content and camera pose information, we follow the established convention of concatenating each image with its corresponding Plücker ray embeddings~\cite{plucker1865newgeometry} along the channel dimension. This concatenated representation is subsequently processed through an MLP-based tokenizer to produce discrete latent tokens. For simplicity, we forego a VAE-based approach and directly tokenize each image into $8 \times 8$ patches. We denote the resulting token sequence from context images and their poses $\{( \textbf{I}^{\text{ctx}}, \bm{\pi}^{\text{ctx}})\}$ as $\mathbf{c} = (c^1, c^2, ..., c^N)$; and similarly define the target token sequence from $\{( \textbf{I}^{\text{tgt}}, \bm{\pi}^{\text{tgt}})\}$ as $\mathbf{x} = (x^1, x^2, ..., x^M)$. Ignoring the facts that the target camera pose are also embedded within  $\mathbf{x}$ and treat them purely as image tokens, we can write the target conditional distribution as $p(\mathbf{x} | \mathbf{c})$.

\begin{figure}[t]
\centering
\includegraphics[width=\textwidth]{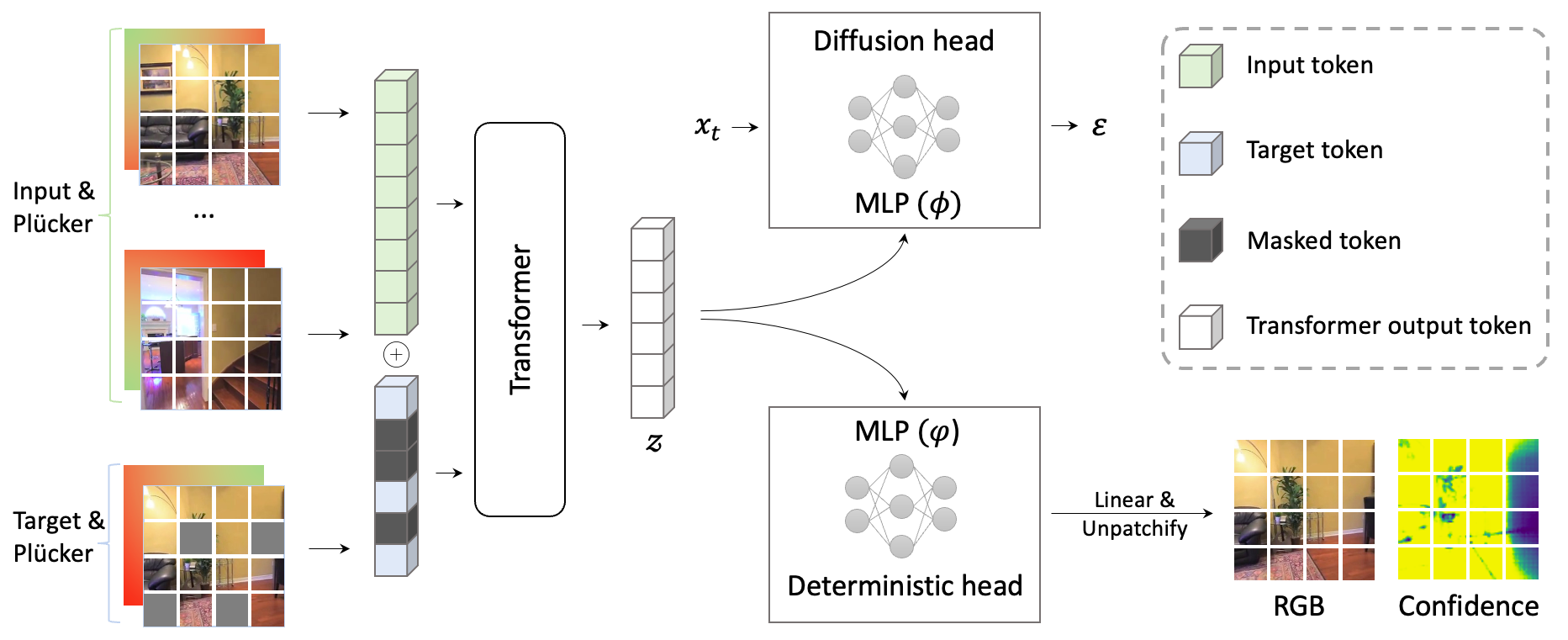}
\caption{\textbf{\mname synthesizes target images from their camera poses and context views (each paired with its Pl\"ucker pose).} During training, we randomly mask the target image, replace masked areas with learnable tokens, and concatenate these with the target's Pl\"ucker embedding. Input views are also tokenized. A Transformer processes both tokenized inputs and the masked target representation to produce a latent $z$. This code inputs to two MLP heads: a deterministic head ($\varphi$) outputs RGB and confidence, while a diffusion head ($\phi$) models the distribution of target tokens conditioned on $\mathbf{z}$. The model is trained end-to-end using a weighted loss combination (Section~\ref{subsec:training-loss}). At inference, the target image is initialized with learned masked tokens for our proposed hybrid sampling (Section~\ref{subsec:hybrid-sampler}).}
\label{fig:framework}
\vspace{-5mm}
\end{figure}

\subsection{Hybrid Masked Autoregressive Models for Novel View Synthesis}
\label{subsec:hybrid-model}

As discussed in Section~\ref{sec:background}, deterministic-based NVS approaches \citep{jin2024lvsmlargeviewsynthesis, sajjadi2022srt} model $p(\mathbf{x} | \mathbf{c})$ to be a deterministic function $F$ of inputs: $p(\mathbf{x} | \mathbf{c}) = \delta(\mathbf{x} - F(\mathbf{c}))$, where $\delta$ is the Dirac delta function. While they have shown strong performance in generating high-fidelity outputs for regions covered by input views, they struggle to handle unseen regions due to their inability to model inherent ambiguity. In contrast, generative models based on diffusion \citep{gao2024cat3d, zhou2025stable, voleti2024sv3d} can generate plausible completions for unobserved regions, but often incur significantly higher computational costs due to iterative sampling over the full image. This trade-off motivates our hybrid design in \mname, which is based on the factorization:
\begin{equation}
    p(\mathbf{x} | \mathbf{c}) = \delta(\mathbf{x}_D - F (\mathbf{c})) \cdot p(\mathbf{x}_S | \mathbf{x}_D, \mathbf{c})
\end{equation}
where \,$\mathbf{x}_D$ and $\mathbf{x}_S$ are disjoint subsets of $\mathbf{x}$ such that $\mathbf{x} = \mathbf{x}_D \cup \mathbf{x}_S$. Intuitively, $\mathbf{x}_D$ corresponds to the tokens that are fully determined by the input context $\mathbf{c}$ (e.g., seen or deterministically visible regions) and can be computed directly as a function $F (\mathbf{c})$. In contrast, $\mathbf{x}_S$ represents tokens in uncertain or unseen regions, which require sampling from a complex conditional distribution $p(\mathbf{x}_S | \mathbf{x}_D, \mathbf{c})$. 
\vspace{-2mm}

\paragraph{Model Architecture} The architecture of \mname is illustrated in the Figure \ref{fig:framework}. \mname  is a masked autoregressive model designed to support both efficient deterministic prediction and flexible stochastic generation by progressively unmasking target tokens.  At the core of our model is a transformer backbone \cite{vaswani2017attention} that extract the target latent representation $\mathbf{z}$ from the partially masked $\mathbf{x}$ and context $\mathbf{c}$. Following previous works \cite{gao2024cat3d, jin2024lvsmlargeviewsynthesis}, we adopt a decoder-only, bi-directional transformer backbone.

%$F_D(\mathbf{c})$
To generate the target tokens, \mname uses two specialized output heads. The \textbf{deterministic head}, parameterized by $\varphi$ in Figure~\ref{fig:framework}, calculates $F(\mathbf{x})$ using the extracted latents $\mathbf{z}$ from the transformer backbone and reconstructs tokens in $\mathbf{x}_D$ in a single forward pass, leveraging regions of high confidence inferred from the context. In contrast, the \textbf{diffusion head}, parameterized by $\phi$ in Figure~\ref{fig:framework}, models the conditional distribution over $\mathbf{x}_S$ and performs iterative denoising to progressively generate plausible content in uncertain or unseen regions. Following MAR~\cite{li2024autoregressive}, both heads are small MLP networks with SiLU activation~\cite{elfwing2017sigmoid} that operate on each token latent individually, and the diffusion head takes an additional time embedding as input. This dual-head design enables \mname to adaptively combine the speed and accuracy of deterministic prediction with the generative capacity of diffusion models, effectively addressing both observed and novel view synthesis scenarios.

In practice, the separation between deterministic and uncertain regions is not known a priori. To address this, we introduce a pixel-wise confidence prediction that estimates an uncertainty score for each pixel. The confidence score of a patch is defined as the minimum confidence among its pixels. Given a threshold $\tau$, patches with confidence above that threshold are assigned to $\mathbf{x}_D$, while the remaining are treated as $\mathbf{x}_S$ and handled via the stochastic generation process. 
\vspace{-2mm}

\subsection{Training Losses}
\label{subsec:training-loss}
We train \mname using a masked autoregressive generative framework~\cite{li2024autoregressive, chang2022maskgit, li2023mage}. At each training step, a binary mask $\mathbf{m}$ is sampled uniformly to mask a subset of the target image patches. Crucially, only the target image is masked (e.g. each selected patch is replaced with a learnable token) while the corresponding target camera pose embeddings are preserved. The model is then optimized to reconstruct the masked patches conditioned on the context and unmasked target information, using a combination of deterministic and diffusion losses:

\paragraph{Deterministic reconstruction loss}
The deterministic head produces token predictions $\hat{\mathbf{x}}$ for masked patches, which yields reconstruction of the target images  $\hat{\mathbf{I}}^{\text{tgt}}$. To supervise this process, we employ standard photometric losses for novel view synthesis, defined as:
\begin{equation}
    \label{eq:l_render}
    \mathcal{L}_{\text{render}} = \text{MSE}(\hat{\mathbf{I}}^{\text{tgt}}, \mathbf{I}^{\text{tgt}}) + \lambda_{\text{p}}  \text{Perceptual}(\hat{\mathbf{I}}^{\text{tgt}}, \mathbf{I}^{\text{tgt}}),
\end{equation}
where $\lambda_p$ is the weight for balancing the perceptual loss~\citep{johnson2016perceptual}. Importantly, this loss is computed over the full image rather than individual patches to encourage spatial consistency.

\paragraph{Confidence loss}
As discussed in Section ~\ref{subsec:hybrid-model}, we augment the deterministic head to output a pixel-level confidence map $\mathbf{s}_p$, where each value indicates the model’s confidence in its prediction. The confidence-aware loss balances the regression error with a regularization term that penalizes overconfidence: 
\begin{equation}
    \label{eq:l_conf}
    \mathcal{L}_{\text{conf}} = \mathbf{m} \odot (\mathbf{s}_p \odot || \hat{\mathbf{I}}^{\text{tgt}} - \mathbf{I}^{\text{tgt}}||_2^2  - \lambda_s \cdot \log \mathbf{s}_p),
\end{equation}
where $\lambda_s$ is a hyper-parameter controlling the regularization term~\cite{wan2018confnet}. The loss is averaged over all masked parts. We compute a patch-wise confidence map $\mathbf{s}$ by taking the minimum value of $\mathbf{s}_p$ within each patch.

\paragraph{Diffusion loss} To model the conditional distribution over uncertain tokens, we incorporate a diffusion model following the formulation of DDPM~\citep{ho2020denoising}. Specifically, we use a linear noise schedule to corrupt the ground truth tokens and train the model to reverse this process. Given a noisy token $x_t$  at timestep $t$ and its corresponding latent $z$ extracted from the transformer backbone, the diffusion head predicts the added noise $\hat{\epsilon}$. The denoising objective is defined over all masked tokens:

\begin{equation}
\label{eq:diff_loss1}
\mathcal{L}_{\text{diff}} = \mathbb{E}_{\epsilon, t} \left[ || \epsilon - \hat{\epsilon} (x_t| t, z)||_2^2 \right],
\end{equation}
where $\epsilon \sim \mathcal{N}(0, I)$ is Gaussian noise, and $t$ is sampled uniformly over the diffusion steps.

To better allocate learning effort, we emphasize uncertain regions during training by predicting a token-wise weighting scheme derived from the patch-level confidence map $\mathbf{s}$. Specifically, we define the weight for each token as $\max(\mathbf{s} , \lambda_d) / \lambda_d$, where $\lambda_d$ is a hyperparameter. This weighting encourages the model to focus more heavily on regions with lower confidence, enhancing generative quality in areas with higher ambiguity.

\paragraph{Total loss} Our model is trained end-to-end using a weighted sum of the aforementioned losses.

\subsection{Hybrid Masked Autoregressive Sampling}
\label{subsec:hybrid-sampler}
\begin{figure}
\centering
\hspace{-1.5em}
\begin{minipage}{0.6\linewidth}{
\includegraphics[width=\textwidth]{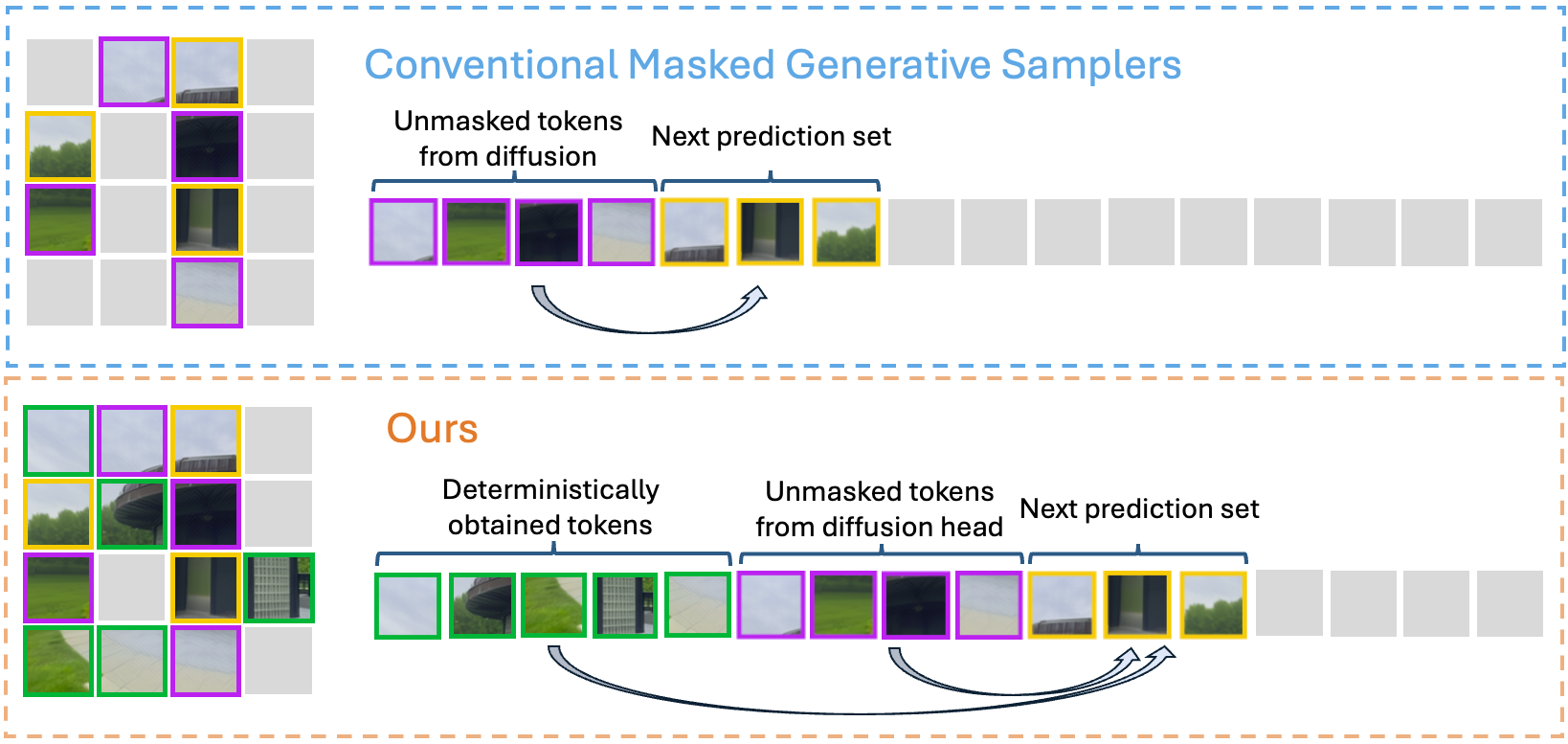}
}\end{minipage}
\hfill
\begin{minipage}{0.4\linewidth}{
\caption{
\textbf{Hybrid Masked Autoregressive Sampler.} Top: Conventional Masked Generative samplers~\cite{li2024autoregressive, chang2022maskgit, li2023mage} predict multiple tokens simultaneously using random ordering.
Bottom (Ours): A deterministic first pass for high confidence tokens, followed by simultaneous random-order sampling for the remaining tokens, significantly boosts rendering times for the NVS task. 
}
\label{fig:sampler}
}\end{minipage}
\vspace{-1em}
\end{figure}
The overall sampling process is illustrated in Figure~\ref{fig:sampler}. Given a masked target image and its corresponding camera pose, \mname performs hybrid inference by first identifying and reconstructing the set of deterministic tokens $\mathbf{x}_D$, and then generating the remaining uncertain tokens $\mathbf{x}_S$ through a diffusion-based process.

In the first stage, the model performs a single forward pass through the deterministic head to predict $\mathbf{x}_D$, guided by the confidence map predicted from the transformer backbone. Tokens with confidence scores greater then a predefined threshold $\tau$ are reconstructed deterministically. 

In the second stage, the remaining masked tokens $\mathbf{x}_S$, are iteratively sampled using the diffusion head. We employ a cosine unmasking schedule following the approach of \cite{li2024autoregressive}, which gradually reveals more tokens at later iterations using $T_S$ unmasking steps. As $|\mathbf{x}_S| \le |\mathbf{x}|$, especially in scenarios where target views significantly overlap with context views, we introduce a dynamic strategy to adjust the number of unmasking steps accordingly. Specifically, given a maximum step budget $T_{\max}$ for unmasking the entire token set $\mathbf{x}$, the number of steps allocated for $\mathbf{x}_S$ is computed using a simple linear scaling rule: $T_S =\left \lceil |\mathbf{x}_S| / |\mathbf{x}| \cdot T_{\max}\right \rceil$. The hyperparameter $T_{\max}$ is fixed across experiments, and $T_S$ is automatically determined by the number of tokens to be unmasked. Despite its simplicity, we find this strategy to be effective in practice and well-suited for varying levels of token uncertainty.
\vspace{-3mm}

\section{Experiments}
\label{sec:experiments}
\paragraph{Datasets} We evaluate \mname on two scene-level novel view synthesis benchmarks: RealEstate10K (CC-BY-4.0)~\cite{zhou2018stereomagnification} and DL3DV (CC-BY-4.0)~\cite{ling2024dl3dv}. RealEstate10K consists of 80K indoor and outdoor video clips sourced from YouTube, while DL3DV features over 10K videos captured across a wide range of real-world locations. We train separate models for each dataset at a resolution of $256 \times 256$. For the RealEstate10K dataset, we adopt the evaluation split from PixelSplat ~\cite{charatan2024pixelsplat}, which primarily features target views located between the $2$ input views, making it suitable for assessing interpolation performance. We refer to this split as \textit{Re10K-2View-Interp}. To evaluate extrapolation ability, we construct a complementary split by swapping the roles of the context and target views, which we denote as \textit{Re10K-2View-Extra}. Additionally, we incorporate the 3-view validation split introduced in Reconfusion~\cite{reconfusion}, labeled as \textit{Re10K-3View}, respectively. For the DL3DV dataset, we follow the validation setup from Zhou et al.~\cite{zhou2025stable}, using the 1-view, 3-view, and 6-view input configurations, which we name \textit{DL3DV-1View}, \textit{DL3DV-3View}, and \textit{DL3DV-6View}, respectively.
\vspace{-2mm}

\paragraph{Experiment Details} Each model is trained for 100K iterations with a batch size of 32, using the AdamW optimizer \cite{loshchilov2017decoupled} with a learning rate of $2 \times 10^{-4}$ and a cosine decay schedule. Training takes approximately two days on $8 \times$ NVIDIA A100 GPUs. During training, we randomly sample 1 or 2 context views and select between 1 and 3 target views per training example. In the main experiments, we report results for predicting a single target view, while results for generating multiple target views are included in the Appendix. To accelerate convergence, we initialize our model using the pretrained transformer backbone from LVSM~\cite{jin2024lvsmlargeviewsynthesis}.  We use a fixed threshold value of $\tau=0.95$ and a maximum sampling steps $T_{\max} = 32$ across our experiments, as we found those values balance well between generation quality and speed. For diffusion sampling, we use $50$ DDPM steps with a CFG value of $2.0$ and a sampling temperature of $0.9$. An details on other hyperparameters of our model are in the Appendix.
\vspace{-2mm}

\paragraph{Baselines} To the best of our knowledge, we are the first method that perform a hybrid render on deterministic and generative method, thus we have no direct competitors. Therefore, we compare our method to different deterministic and generative baselines. For deterministic methods, we compare \mname to MVSplat \cite{chen2024mvsplat}, LVSM \cite{jin2024lvsmlargeviewsynthesis}. For generative approaches, we compare \mname with ViewCrafter \cite{yu2024viewcrafter} and SEVA \cite{zhou2025stable}.
\vspace{-2mm}
\subsection{Experiment results}

\begin{table*}[t]
\centering
\caption{Quantitative results on RealEstate10K across different validation splits. Best results are \cellcolor{best}highlighted in red, second-best in \cellcolor{second}orange.}
\resizebox{\textwidth}{!}{%
\begin{tabular}{l|c|ccc|ccc|ccc}
\toprule
\multirow{2}{*}{Method} & \multirow{2}{*}{Params (M)} & \multicolumn{3}{c|}{Re10K-2view-extra} & \multicolumn{3}{c|}{Re10K-2view-interp} & \multicolumn{3}{c}{Re10K-3view} \\
& & PSNR $\uparrow$ & LPIPS $\downarrow$ & SSIM $\uparrow$ & PSNR $\uparrow$ & LPIPS $\downarrow$ & SSIM $\uparrow$ & PSNR $\uparrow$ & LPIPS $\downarrow$ & SSIM $\uparrow$ \\
\midrule
\multicolumn{11}{l}{\textbf{Deterministic}} \\
MVSplat \cite{chen2024mvsplat}& 12.0 & 23.30& 0.160& 0.830& 26.39 & 0.128 & 0.869 & 25.64 & 0.142 & 0.857 \\
DepthSplat \cite{xu2024depthsplat} & 360 & 24.57& 0.158& 0.848& 27.44 & 0.119 & 0.887& 22.54 & 0.177 & 0.824 \\
LVSM \cite{jin2024lvsmlargeviewsynthesis} & 171 & \cellcolor{second}28.51& 0.117& \cellcolor{second}0.882& \cellcolor{best}29.67 & \cellcolor{second}0.098 & \cellcolor{best}0.906 & \cellcolor{second}30.04 & 0.090 & \cellcolor{second}0.936 \\
\midrule
\multicolumn{11}{l}{\textbf{Diffusion-based}} \\
ViewCrafter \cite{yu2024viewcrafter} & N/A & -& -& -& 21.42 & 0.203 & 0.710 & 22.81 & 0.164 & 0.830 \\
SEVA \cite{zhou2025stable} & 1300 & 24.00& \cellcolor{best}0.100& 0.797& 25.66 & \cellcolor{best}0.061 & 0.847 & 27.57 & \cellcolor{best}0.073 & 0.892 \\
\midrule
\mname & 271& \cellcolor{best}28.95& \cellcolor{second}0.107& \cellcolor{best}0.897& \cellcolor{second}28.85 & 0.101 & \cellcolor{second}0.899 & \cellcolor{best}31.06 & \cellcolor{second}0.084 & \cellcolor{best}0.946 \\
\bottomrule
\end{tabular}
}
\label{tab:re10k_all}
\end{table*}
\begin{table*}[t]
\centering
\caption{Quantitative results on DL3DV across 1-view, 3-view, and 6-view settings. Best results are \cellcolor{best}highlighted in red, second-best in \cellcolor{second}orange.}
\resizebox{\textwidth}{!}{%
\begin{tabular}{l|ccc|ccc|ccc}
\toprule
\multirow{2}{*}{Method} & \multicolumn{3}{c|}{DL3DV-1view} & \multicolumn{3}{c|}{DL3DV-3view} & \multicolumn{3}{c}{DL3DV-6view} \\
 & PSNR $\uparrow$ & LPIPS $\downarrow$ & SSIM $\uparrow$ & PSNR $\uparrow$ & LPIPS $\downarrow$ & SSIM $\uparrow$ & PSNR $\uparrow$ & LPIPS $\downarrow$ & SSIM $\uparrow$ \\
\midrule
\multicolumn{10}{l}{\textbf{Deterministic}} \\
DepthSplat \cite{xu2024depthsplat}  & 9.63 & 0.580 & \cellcolor{second}0.349& 12.52 & 0.405 & \cellcolor{second}0.452& 15.72 & 0.481 & \cellcolor{second}0.513\\
\midrule
\multicolumn{10}{l}{\textbf{Diffusion-based}} \\
ViewCrafter \cite{yu2024viewcrafter} & 8.97 & 0.616 & 0.323 & 11.50 & 0.576 & 0.400 & 13.78 & 0.558 & 0.469 \\
SEVA \cite{zhou2025stable}       & \cellcolor{best}13.01& \cellcolor{best}0.484 & \cellcolor{best}0.360 & \cellcolor{second}15.95& \cellcolor{best}0.316 & \cellcolor{best}0.480& \cellcolor{best}17.98 & \cellcolor{best}0.232 & \cellcolor{best}0.546 \\
\midrule
\mname         & \cellcolor{second}12.81& \cellcolor{second}0.574& 0.269& \cellcolor{best}16.37& \cellcolor{second}0.386& 0.444& \cellcolor{second}17.33& \cellcolor{second}0.326& 0.476\\
\bottomrule
\end{tabular}
}
\label{tab:dl3dv_results}
\end{table*}

\begin{figure}[!ht]
\centering
\includegraphics[width=0.95\textwidth]{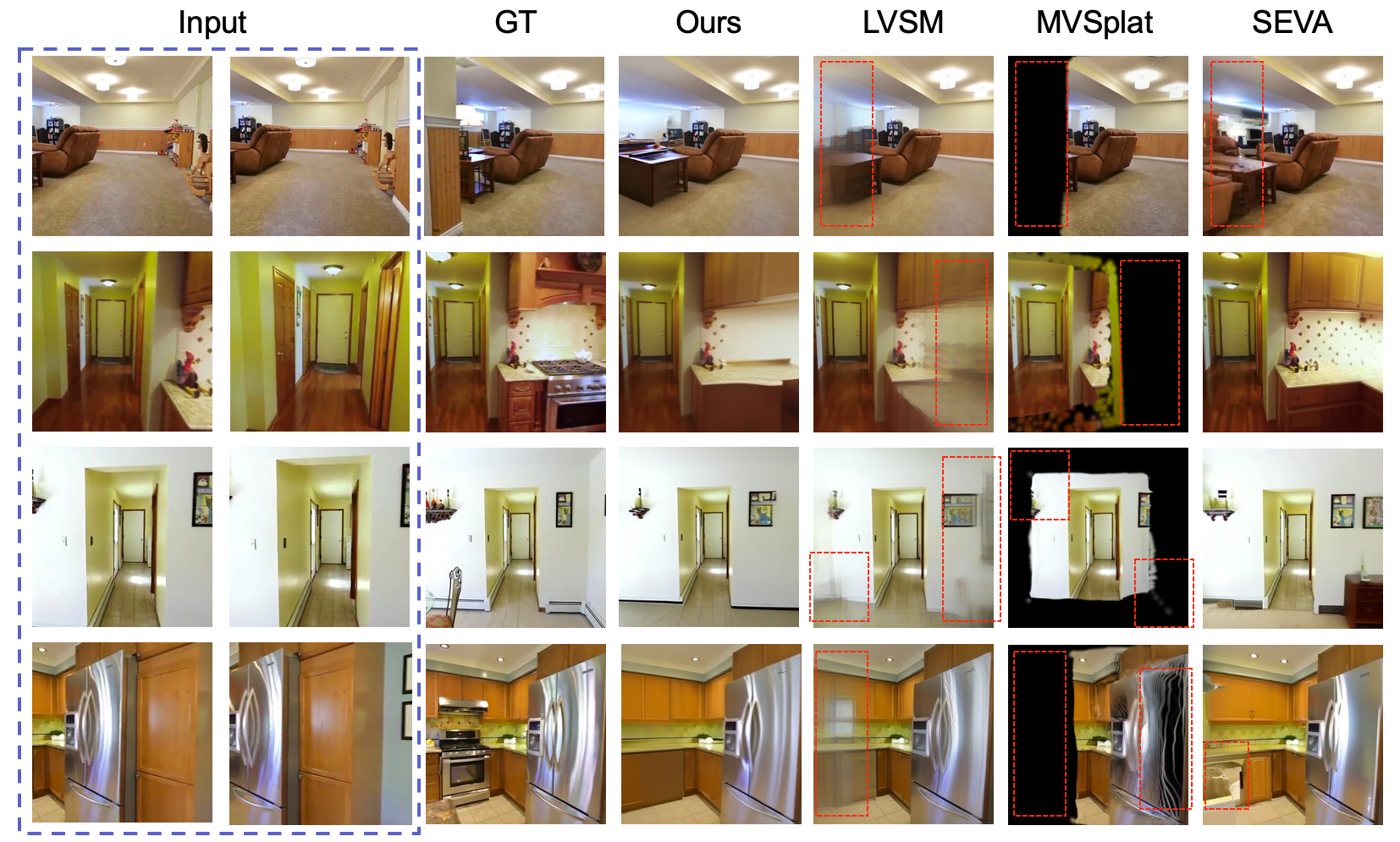}
\caption{\textbf{Qualitative results on Re10K} Evaluation of \mname on the challenging \textit{Re10K-2View-Extra} extrapolation set, comparing it with LVSM~\cite{jin2024lvsmlargeviewsynthesis}, MVSplat~\cite{chen2024mvsplat}, and SEVA~\cite{zhou2025stable}. \mname not only renders sharp details in observed regions but also generates plausible content for unseen areas. More results can be viewed in the Appendix.}
\label{fig:qualitative-re10k}
\vspace{-7mm}
\end{figure}
\vspace{-1mm}
\paragraph{Quantitative results} Tables \ref{tab:re10k_all} and \ref{tab:dl3dv_results} present a comprehensive comparison of our method \mname against both deterministic and diffusion-based baselines on the RealEstate10K and DL3DV datasets, respectively. On RealEstate10K, \mname consistently achieves top-tier performance across all splits.
On the Re10K-2view-interp split, it matches LVSM closely, trailing by only 0.82 PSNR (28.85 vs. 29.67).
On the Re10K-2view-extra split, it obtains the highest PSNR (28.95) and SSIM (0.897), outperforming the second-best LVSM by 0.44 PSNR and 0.015 SSIM, while maintaining a second-best LPIPS of 0.107; showing its capabilities of doing extrapolation.  On the Re10K-3view setting, \mname surpasses all baselines with the best PSNR (31.06) and SSIM (0.946), while having the second best LPIPS of 0.084.

On the DL3DV benchmark, \mname delivers competitive performance across all input configurations. It achieves the second-best PSNR in both the 1-view (12.81) and 6-view (17.33) settings, and ranks second in LPIPS across all three input setups. Notably, in the 3-view setting, \mname achieves the highest PSNR among all methods. While its SSIM lags behind DepthSplat, \mname outperforms ViewCrafter in both the 3-view and 6-view scenarios. These results underscore the robustness and adaptability of our hybrid framework, demonstrating its effectiveness in handling diverse view configurations and maintaining a strong balance between reconstruction fidelity and perceptual quality.
\vspace{-3mm}

\paragraph{Varying overlap ratios} Following NoPoSplat~\citep{ye2024no}, we evaluate \mname on Re10k test set with varying camera overlaps based on ratio of image overlap: small ($0.05\%$-$0.3\%$), medium ($0.3\%$ - $0.55\%$), and large ($0.55\%$ - $0.8\%$), determined using dense feature matching method, RoMA~\citep{edstedt2024roma}. The results, shown in the Table~\ref{tab:overlap}, demonstrate that \mname outperforms NoPoSplat on all metrics and validation sets. As with NoPoSplat, we also observed that performance improves as the overlap ratio increases, which suggests less scene occlusion. Notably, our method remains robust even with a small camera overlap, outperforming NoPoSplat across all metrics. This confirms \mname's effectiveness across various datasets and overlap ratios.

\begin{table*}[!h]
\centering
\caption{Quantitative comparison across overlaping ratios (Small, Medium, Large). Best results are \cellcolor{best}highlighted in red, second-best in \cellcolor{second}orange.}
\resizebox{\textwidth}{!}{%
\begin{tabular}{l|ccc|ccc|ccc}
\toprule
\multirow{2}{*}{Method} & \multicolumn{3}{c|}{Small} & \multicolumn{3}{c|}{Medium} & \multicolumn{3}{c}{Large} \\
 & PSNR $\uparrow$ & SSIM $\uparrow$ & LPIPS $\downarrow$ & PSNR $\uparrow$ & SSIM $\uparrow$ & LPIPS $\downarrow$ & PSNR $\uparrow$ & SSIM $\uparrow$ & LPIPS $\downarrow$ \\
\midrule
NoPoSplat & \cellcolor{second}22.514 & \cellcolor{second}0.784 & \cellcolor{second}0.210 & \cellcolor{second}24.899 & \cellcolor{second}0.839 & \cellcolor{second}0.160 & \cellcolor{second}27.411 & \cellcolor{second}0.883 & \cellcolor{second}0.119 \\
Ours       & \cellcolor{best}23.558 & \cellcolor{best}0.806 & \cellcolor{best}0.176 & \cellcolor{best}26.713 & \cellcolor{best}0.862 & \cellcolor{best}0.130 & \cellcolor{best}29.782 & \cellcolor{best}0.907 & \cellcolor{best}0.094 \\
\bottomrule
\end{tabular}
}
\label{tab:overlap}
\end{table*}

\paragraph{Qualitative results} presented in Figure~\ref{fig:qualitative-re10k}, highlight the performance of our method against several methods. Firstly, MVSplat~\cite{chen2024mvsplat}, as a deterministic method employing 3D Gaussians, is unable to generate content beyond the provided context images, resulting in black rendered areas in unobserved regions. Similarly, while LVSM~\cite{jin2024lvsmlargeviewsynthesis} avoids such black areas by forgoing 3D inductive biases, its non-generative nature results in blurry predictions for unseen pixels. Our method overcomes these limitations of deterministic approaches, demonstrating the ability to both accurately render observed regions and plausibly generate content in unobserved areas. Finally, in comparison to SEVA~\cite{zhou2025stable}, a considerably larger model with 1.3B parameters (versus our 271M parameters), our approach achieves comparable performance on visible regions and produces results with fewer artifacts.

\subsection{Ablation study and Analysis}
\vspace{-5mm}
\begin{table}[ht]
\setlength{\tabcolsep}{4pt} 
\centering

\begin{minipage}{0.49\linewidth}
\centering
% \begin{tabular}{|c|c|}
% \hline
% A & B \\
% \hline
% 1 & 2 \\
% \hline
% \end{tabular}
% \caption{Table 1}

\caption{Ablation study on threshold $\tau$. Higher $\tau$ improves image quality but increases transformer calls and runtime.}
\label{tab:ablation} % Label for the table
\begin{tabular}{cccc}
\toprule % Using booktabs style, consistent with template's rules
  {$\tau$} & {Time (s)} & {\# Trans. Calls} & {LPIPS} $\downarrow$ \\
\midrule
  0    & 0.02 &  1.00 & 0.398 \\
  0.5  & 2.71 & 12.31 & 0.394 \\
  0.8  & 4.30 & 18.99 & 0.389 \\
  0.9  & 4.62 & 20.30 & 0.387 \\
  0.95 & 4.77 & 21.14 & 0.386 \\
  1    & 7.63 & 33.00 & 0.377 \\
\bottomrule
\end{tabular}

\end{minipage}
\hfill
\begin{minipage}{0.49\linewidth}
\centering
    \caption{Ablation study on the number of context views. Increasing the number of context views ($N_c$) improves image synthesis quality by providing more deterministic tokens and reducing the number of average transformer calls, due to higher confidence in a larger portion of the scene.}
    \label{tab:ablation2} 
    \begin{tabular}{cccc}
      \toprule
      \textbf{$N_c$} & {\# Deter Tokens} & {\# Trans. Calls} & {LPIPS} $\downarrow$ \\
      \midrule
      1 & 119.24 & 29.55 & 0.574 \\
      3 & 394.68 & 21.14 & 0.386 \\
      6 & 527.25 & 16.98 & 0.326 \\
      \bottomrule
    \end{tabular}
\end{minipage}

\end{table}
\vspace{-3mm}
\paragraph{Effect on threshold $\tau$ during sampling}
We control the balance between \mname's deterministic and diffusion heads during sampling using a threshold $\tau$ in Table~\ref{tab:ablation}. Setting $\tau=0$ engages only the deterministic head, enabling \mname to predict all tokens in $0.02$s with one transformer call. Conversely, setting $\tau=1$ relies exclusively on the diffusion head for sampling target views, which yields the optimal LPIPS score in our experiments. We observe that incrementally increasing $\tau$ from $0$ to $1$ enhances LPIPS performance, though at the cost of increased runtime due to more frequent transformer and diffusion sampling operations. Thus, by adjusting $\tau$, our dual-head model offers a flexible mechanism to trade off inference speed against generative quality.

\vspace{-0.5em}

\paragraph{Effect on the number of context views}
We conduct an ablation study on the effect of varying the number of context views, as shown in \ref{tab:ablation2}. Results indicate that increasing the number of input views leads to a significant boost in mean LPIPS. Furthermore, with more context available, the model exhibits higher confidence, resulting in a greater proportion of tokens being handled deterministically. This, in turn, reduces the number of transformer calls required during sampling, leading to improved computational efficiency.
\vspace{-0.5em}
\paragraph{Run time analysis}
Unlike conventional methods with fixed rendering times, \mname offers operational flexibility by adaptively engaging its deterministic and generative heads. For instance, \mname renders an image in approximately $5$s when $\tau=0.95$ (details in Table~\ref{tab:ablation}). This is considerably faster than generative counterparts like SEVA~\cite{zhou2025stable}, which takes about $1$ minute to sample an image. While purely deterministic methods~\cite{chen2024mvsplat} achieve sub-second rendering, they sacrifice the ability to generate content for unobserved target regions. \mname thus provides a compelling trade-off: it achieves strong generative capabilities for a modest increase in runtime compared to deterministic approaches, while remaining significantly more efficient than other generative models.
\vspace{-2mm}

\begin{table*}[t]
\centering
\caption{Ablation results on DL3DV across 1-view, 3-view, and 6-view settings comparing pretrained and random initialization. Best results are \cellcolor{best}highlighted in red, second-best in \cellcolor{second}orange.}
\resizebox{\textwidth}{!}{%
\begin{tabular}{l|ccc|ccc|ccc}
\toprule
\multirow{2}{*}{Method} & \multicolumn{3}{c|}{DL3DV-1view} & \multicolumn{3}{c|}{DL3DV-3view} & \multicolumn{3}{c}{DL3DV-6view} \\
 & PSNR $\uparrow$ & LPIPS $\downarrow$ & SSIM $\uparrow$ & PSNR $\uparrow$ & LPIPS $\downarrow$ & SSIM $\uparrow$ & PSNR $\uparrow$ & LPIPS $\downarrow$ & SSIM $\uparrow$ \\
\midrule
Pretrained & \cellcolor{best}12.81 & \cellcolor{second}0.574 & \cellcolor{best}0.269 & \cellcolor{best}16.37 & \cellcolor{second}0.386 & \cellcolor{best}0.444 & \cellcolor{best}17.33 & \cellcolor{second}0.326 & \cellcolor{best}0.476 \\
Random Init. & \cellcolor{second}11.80 & \cellcolor{best}0.543 & \cellcolor{second}0.256 & \cellcolor{second}14.46 & \cellcolor{best}0.374 & \cellcolor{second}0.370 & \cellcolor{second}15.43 & \cellcolor{best}0.318 & \cellcolor{second}0.404 \\
\bottomrule
\end{tabular}
}
\label{tab:random-weight}
\end{table*}
\paragraph{Backbone initialization}
As mentioned in Section~\ref{sec:experiments}, we initialize our model using pretrained weight from LVSM~\cite{jin2024lvsmlargeviewsynthesis}. To further demonstrate the strength of our method, we train the model from scratch on DL3DV with randomly initialized weights, without relying on LVSM pretrained on Re10K. This variant shares the same settings as the pretrained version, except for a larger batch size (512 vs. 32) to stabilize training. The results, presented in the Table~\ref{tab:random-weight}, reveal that even without pretraining, our model performs comparably—showing slightly lower PSNR and SSIM but improved LPIPS. Remarkably, even with random initialization, our method consistently outperforms ViewCrafter and DepthSplat, underscoring that our performance stems from the strength of our hybrid deterministic-generative design, rather than dependence on LVSM initialization and Re10K pretraining.
\section{Discussion and Conclusion}
\paragraph{Limitations and Future Work}
While our method achieves competitive performance, it also has several limitations. First, because we operate directly in pixel space, each image is represented by a large number of tokens (e.g., $32 \times 32 = 1024$), which increases memory and computational requirements. A promising direction for future work is to adapt our framework to operate in the latent space of a pretrained VAE~\cite{kingma2013auto}, which would reduce the token count while preserving semantic content. Second, unlike recent diffusion-based NVS approaches~\cite{gao2024cat3d, zhou2025stable, yu2024viewcrafter}, our model does not make use of any pretrained text-to-image priors. Integrating such powerful generative priors \citep{li2024autoregressive, deng2024autoregressive} could enhance the model’s ability to hallucinate plausible unseen regions and improve visual fidelity in sparse-view settings. We also leave for future exploration techniques to further accelerate sampling and incorporate temporal consistency for video-based novel view synthesis. 
On the social impact side, this work could enable deepfake information, so users will be required to follow usage guidelines.

\vspace{-0.5em}

\paragraph{Conclusion} We have presented \mname, a hybrid framework for novel view synthesis that unifies deterministic and generative modeling to handle both seen and unseen regions effectively. By leveraging a confidence-aware mechanism, our model adaptively allocates computation between a fast deterministic head and a diffusion-based head, achieving a strong balance between efficiency and image quality. Extensive experiments on RealEstate10K and DL3DV demonstrate that \mname is competitive with both deterministic and diffusion-only baselines across various input configurations. Our results suggest a promising direction for designing more efficient approaches to novel view synthesis.

\section*{Acknowledgements}
Stephan Mandt acknowledges funding from the National Science Foundation (NSF) through an NSF CAREER Award IIS-2047418, IIS-2007719, the NSF LEAP Center, and the Hasso Plattner Research Center at UCI. Xiaohui Xie acknowledges funding from NIH 1P01CA288662-01A1 and Kay Family Foundation. Parts of this research were supported by the Intelligence Advanced Research Projects Activity (IARPA) via the Department of Interior/ Interior Business Center (DOI/IBC) contract number 140D0423C0075. The U.S. Government is authorized to reproduce and distribute reprints for Governmental purposes
notwithstanding any copyright annotation thereon. Disclaimer: The views and conclusions
contained herein are those of the authors and should not be interpreted as necessarily
representing the official policies or endorsements, either expressed or implied, of IARPA, DOI/IBC, or the U.S. Government.

\clearpage
\bibliography{bib}

@inproceedings{barron2021mip,
  title={Mip-nerf: A multiscale representation for anti-aliasing neural radiance fields},
  author={Barron, Jonathan T and Mildenhall, Ben and Tancik, Matthew and Hedman, Peter and Martin-Brualla, Ricardo and Srinivasan, Pratul P},
  booktitle={Proceedings of the IEEE/CVF international conference on computer vision},
  pages={5855--5864},
  year={2021}
}

@inproceedings{barron2023zip,
  title={Zip-nerf: Anti-aliased grid-based neural radiance fields},
  author={Barron, Jonathan T and Mildenhall, Ben and Verbin, Dor and Srinivasan, Pratul P and Hedman, Peter},
  booktitle={Proceedings of the IEEE/CVF International Conference on Computer Vision},
  pages={19697--19705},
  year={2023}
}

@inproceedings{edstedt2024roma,
  title={Roma: Robust dense feature matching},
  author={Edstedt, Johan and Sun, Qiyu and B{\"o}kman, Georg and Wadenb{\"a}ck, M{\aa}rten and Felsberg, Michael},
  booktitle={Proceedings of the IEEE/CVF Conference on Computer Vision and Pattern Recognition},
  pages={19790--19800},
  year={2024}
}

@inproceedings{chan2022efficient,
  title={Efficient geometry-aware 3d generative adversarial networks},
  author={Chan, Eric R and Lin, Connor Z and Chan, Matthew A and Nagano, Koki and Pan, Boxiao and De Mello, Shalini and Gallo, Orazio and Guibas, Leonidas J and Tremblay, Jonathan and Khamis, Sameh and others},
  booktitle={Proceedings of the IEEE/CVF conference on computer vision and pattern recognition},
  pages={16123--16133},
  year={2022}
}

@inproceedings{charatan2024pixelsplat,
  title={pixelsplat: 3d gaussian splats from image pairs for scalable generalizable 3d reconstruction},
  author={Charatan, David and Li, Sizhe Lester and Tagliasacchi, Andrea and Sitzmann, Vincent},
  booktitle={Proceedings of the IEEE/CVF conference on computer vision and pattern recognition},
  pages={19457--19467},
  year={2024}
}

@article{chaurasia2013depth,
  title={Depth synthesis and local warps for plausible image-based navigation},
  author={Chaurasia, Gaurav and Duchene, Sylvain and Sorkine-Hornung, Olga and Drettakis, George},
  journal={ACM transactions on graphics (TOG)},
  volume={32},
  number={3},
  pages={1--12},
  year={2013},
  publisher={ACM New York, NY, USA}
}

@inproceedings{chen2022tensorf,
  title={Tensorf: Tensorial radiance fields},
  author={Chen, Anpei and Xu, Zexiang and Geiger, Andreas and Yu, Jingyi and Su, Hao},
  booktitle={European conference on computer vision},
  pages={333--350},
  year={2022},
  organization={Springer}
}

@article{chen2023factor,
  title={Factor fields: A unified framework for neural fields and beyond},
  author={Chen, Anpei and Xu, Zexiang and Wei, Xinyue and Tang, Siyu and Su, Hao and Geiger, Andreas},
  journal={arXiv preprint arXiv:2302.01226},
  year={2023}
}

@inproceedings{chen2024mvsplat,
  title={Mvsplat: Efficient 3d gaussian splatting from sparse multi-view images},
  author={Chen, Yuedong and Xu, Haofei and Zheng, Chuanxia and Zhuang, Bohan and Pollefeys, Marc and Geiger, Andreas and Cham, Tat-Jen and Cai, Jianfei},
  booktitle={European Conference on Computer Vision},
  pages={370--386},
  year={2024},
  organization={Springer}
}

@inproceedings{choi2019extreme,
  title={Extreme view synthesis},
  author={Choi, Inchang and Gallo, Orazio and Troccoli, Alejandro and Kim, Min H and Kautz, Jan},
  booktitle={Proceedings of the IEEE/CVF International Conference on Computer Vision},
  pages={7781--7790},
  year={2019}
}

@inproceedings{davis2012unstructured,
  title={Unstructured light fields},
  author={Davis, Abe and Levoy, Marc and Durand, Fredo},
  booktitle={Computer Graphics Forum},
  volume={31},
  number={2pt1},
  pages={305--314},
  year={2012},
  organization={Wiley Online Library}
}

@incollection{debevec2023modeling,
  title={Modeling and rendering architecture from photographs: A hybrid geometry-and image-based approach},
  author={Debevec, Paul E and Taylor, Camillo J and Malik, Jitendra},
  booktitle={Seminal Graphics Papers: Pushing the Boundaries, Volume 2},
  pages={465--474},
  year={2023}
}

@inproceedings{deitke2023objaverse,
  author    = {Deitke, Matt and Schwenk, Dustin and Salvador, Jordi and Weihs, Luca and Michel, Oscar and VanderBilt, Eli and Schmidt, Ludwig and Ehsani, Kiana and Kembhavi, Aniruddha and Farhadi, Ali},
  title     = {Objaverse: A Universe of Annotated 3D Objects},
  booktitle = {Proceedings of the IEEE/CVF Conference on Computer Vision and Pattern Recognition (CVPR)},
  year      = {2023},
  pages     = {13142--13153}
}

@inproceedings{feng2022neural,
  title={Neural points: Point cloud representation with neural fields for arbitrary upsampling},
  author={Feng, Wanquan and Li, Jin and Cai, Hongrui and Luo, Xiaonan and Zhang, Juyong},
  booktitle={Proceedings of the IEEE/CVF conference on computer vision and pattern recognition},
  pages={18633--18642},
  year={2022}
}

@inproceedings{fridovich2022plenoxels,
  title={Plenoxels: Radiance fields without neural networks},
  author={Fridovich-Keil, Sara and Yu, Alex and Tancik, Matthew and Chen, Qinhong and Recht, Benjamin and Kanazawa, Angjoo},
  booktitle={Proceedings of the IEEE/CVF conference on computer vision and pattern recognition},
  pages={5501--5510},
  year={2022}
}

@article{gao2024cat3d,
  title={Cat3d: Create anything in 3d with multi-view diffusion models},
  author={Gao, Ruiqi and Holynski, Aleksander and Henzler, Philipp and Brussee, Arthur and Martin-Brualla, Ricardo and Srinivasan, Pratul and Barron, Jonathan T and Poole, Ben},
  journal={Advances in Neural Information Processing Systems},
  year={2024}
}

@misc{he2024cameractrl,
  author       = {He, Hao and Xu, Yinghao and Guo, Yuwei and Wetzstein, Gordon and Dai, Bo and Li, Hongsheng and Yang, Ceyuan},
  title        = {CameraCtrl: Enabling Camera Control for Text{-}to{-}Video Generation},
  year         = {2024},
  eprint       = {2404.02101},
  archivePrefix= {arXiv},
  primaryClass = {cs.CV}
}

@inproceedings{hedman2021baking,
  author    = {Hedman, Peter and Srinivasan, Pratul P. and Mildenhall, Ben and Barron, Jonathan T. and Debevec, Paul},
  title     = {Baking Neural Radiance Fields for Real{-}Time View Synthesis},
  booktitle = {Proceedings of the IEEE/CVF International Conference on Computer Vision (ICCV)},
  year      = {2021},
  pages     = {5875--5884}
}

@inproceedings{heigl1999plenoptic,
  author    = {Heigl, Benno and Koch, Reinhard and Pollefeys, Marc and Denzler, Joachim and Van Gool, Luc},
  title     = {Plenoptic Modeling and Rendering from Image Sequences Taken by a Hand{-}Held Camera},
  booktitle = {Mustererkennung 1999: 21st DAGM Symposium},
  year      = {1999},
  pages     = {94--101}
}

@misc{hong2024lrmlargereconstructionmodel,
  author       = {Hong, Yicong and Zhang, Kai and Gu, Jiuxiang and Bi, Sai and Zhou, Yang and Liu, Difan and Liu, Feng and Sunkavalli, Kalyan and Bui, Trung and Tan, Hao},
  title        = {{LRM}: Large Reconstruction Model for Single Image to 3D},
  year         = {2024},
  eprint       = {2311.04400},
  archivePrefix= {arXiv},
  primaryClass = {cs.CV}
}

@inproceedings{jancosek2011multi,
  author    = {Jancosek, Michal and Pajdla, Tomas},
  title     = {Multi{-}View Reconstruction Preserving Weakly{-}Supported Surfaces},
  booktitle = {Proceedings of the IEEE Conference on Computer Vision and Pattern Recognition (CVPR)},
  year      = {2011},
  pages     = {3121--3128}
}

@misc{jiang2023leap,
  author       = {Jiang, Hanwen and Jiang, Zhenyu and Zhao, Yue and Huang, Qixing},
  title        = {LEAP: Liberate Sparse{-}View 3D Modeling from Camera Poses},
  year         = {2023},
  eprint       = {2310.01410},
  archivePrefix= {arXiv},
  primaryClass = {cs.CV}
}

@inproceedings{jiang2024forge,
  author    = {Jiang, Hanwen and Jiang, Zhenyu and Grauman, Kristen and Zhu, Yuke},
  title     = {Few{-}View Object Reconstruction with Unknown Categories and Camera Poses},
  booktitle = {Proceedings of the International Conference on 3D Vision (3DV)},
  year      = {2024},
  pages     = {31--41}
}

@misc{johari2022geonerfgeneralizingnerfgeometry,
  author       = {Johari, Mohammad Mahdi and Lepoittevin, Yann and Fleuret, François},
  title        = {GeoNeRF: Generalizing NeRF with Geometry Priors},
  year         = {2022},
  eprint       = {2111.13539},
  archivePrefix= {arXiv},
  primaryClass = {cs.CV}
}

@inproceedings{johnson2016perceptual,
  author    = {Johnson, Justin and Alahi, Alexandre and Fei{-}Fei, Li},
  title     = {Perceptual Losses for Real{-}Time Style Transfer and Super{-}Resolution},
  booktitle = {Proceedings of the European Conference on Computer Vision (ECCV)},
  year      = {2016},
  pages     = {694--711}
}

@article{kerbl3Dgaussians,
  author  = {Kerbl, Bernhard and Kopanas, Georgios and Leimk{\"u}hler, Thomas and Drettakis, George},
  title   = {3D Gaussian Splatting for Real{-}Time Radiance Field Rendering},
  journal = {ACM Transactions on Graphics},
  volume  = {42},
  number  = {4},
  year    = {2023},
  month   = {July}
}

@misc{kong2024eschernet,
  author       = {Kong, Xin and Liu, Shikun and Lyu, Xiaoyang and Taher, Marwan and Qi, Xiaojuan and Davison, Andrew J.},
  title        = {EscherNet: A Generative Model for Scalable View Synthesis},
  year         = {2024},
  eprint       = {2402.03908},
  archivePrefix= {arXiv},
  primaryClass = {cs.CV}
}

@inproceedings{kulhanek2022viewformer,
  author    = {Kulh{\'{a}}nek, Jon{\'{a}}{\v{s}} and Derner, Erik and Sattler, Torsten and Babu{\v{s}}ka, Robert},
  title     = {Viewformer: NeRF{-}Free Neural Rendering from Few Images Using Transformers},
  booktitle = {European Conference on Computer Vision (ECCV)},
  year      = {2022}
}

@misc{li2023instant3d,
  author       = {Li, Jiahao and Tan, Hao and Zhang, Kai and Xu, Zexiang and Luan, Fujun and Xu, Yinghao and Hong, Yicong and Sunkavalli, Kalyan and Shakhnarovich, Greg and Bi, Sai},
  title        = {Instant3D: Fast Text{-}to{-}3D with Sparse{-}View Generation and Large Reconstruction Model},
  year         = {2023},
  eprint       = {2311.06214},
  archivePrefix= {arXiv},
  primaryClass = {cs.CV}
}

@inproceedings{liu2020nsvf,
  author    = {Liu, Lingjie and Gu, Jiatao and Lin, Kyaw Zaw and Chua, Tat{-}Seng and Theobalt, Christian},
  title     = {Neural Sparse Voxel Fields},
  booktitle = {Advances in Neural Information Processing Systems (NeurIPS)},
  volume    = {33},
  pages     = {15651--15663},
  year      = {2020}
}

@misc{liu2023zero1to3,
  author       = {Liu, Ruoshi and Wu, Rundi and Van Hoorick, Basile and Tokmakov, Pavel and Zakharov, Sergey and Vondrick, Carl},
  title        = {Zero{-}1{-}to{-}3: Zero{-}Shot One Image to 3D Object},
  year         = {2023},
  eprint       = {2302.00000},
  archivePrefix= {arXiv},
  primaryClass = {cs.CV}
}

@inproceedings{liu2022neuralray,
  author    = {Liu, Yuan and Peng, Sida and Liu, Lingjie and Wang, Qianqian and Wang, Peng and Theobalt, Christian and Zhou, Xiaowei and Wang, Wenping},
  title     = {Neural Rays for Occlusion{-}Aware Image{-}Based Rendering},
  booktitle = {Proceedings of the IEEE/CVF Conference on Computer Vision and Pattern Recognition (CVPR)},
  year      = {2022},
  pages     = {7824--7833}
}

@misc{liu2023syncdreamer,
  author       = {Liu, Yuan and Lin, Cheng and Zeng, Zijiao and Long, Xiaoxiao and Liu, Lingjie and Komura, Taku and Wang, Wenping},
  title        = {SyncDreamer: Generating Multiview{-}Consistent Images from a Single{-}View Image},
  year         = {2023},
  eprint       = {2309.03453},
  archivePrefix= {arXiv},
  primaryClass = {cs.CV}
}

@misc{long2023wonder3d,
  author       = {Long, Xiaoxiao and Guo, Yuan{-}Chen and Lin, Cheng and Liu, Yuan and Dou, Zhiyang and Liu, Lingjie and Ma, Yuexin and Zhang, Song{-}Hai and Habermann, Marc and Theobalt, Christian and others},
  title        = {Wonder3D: Single Image to 3D Using Cross{-}Domain Diffusion},
  year         = {2023},
  eprint       = {2310.15008},
  archivePrefix= {arXiv},
  primaryClass = {cs.CV}
}

@inproceedings{niemeyer2022regnerf,
  author    = {Niemeyer, Michael and Barron, Jonathan T. and Mildenhall, Ben and Sajjadi, Mehdi SM. and Geiger, Andreas and Radwan, Noha},
  title     = {RegNeRF: Regularizing Neural Radiance Fields for View Synthesis from Sparse Inputs},
  booktitle = {Proceedings of the IEEE/CVF Conference on Computer Vision and Pattern Recognition (CVPR)},
  year      = {2022},
  pages     = {5480--5490}
}

@article{penner2017soft,
  author  = {Penner, Eric and Zhang, Li},
  title   = {Soft 3D Reconstruction for View Synthesis},
  journal = {ACM Transactions on Graphics},
  volume  = {36},
  number  = {6},
  pages   = {1--11},
  year    = {2017}
}

@article{plucker1865newgeometry,
  author  = {Pl{\"u}cker, Julius},
  title   = {On a New Geometry of Space},
  journal = {Philosophical Transactions of the Royal Society of London},
  pages   = {725--791},
  year    = {1865}
}

@misc{ramesh2022hierarchical,
  author       = {Ramesh, Aditya and Dhariwal, Prafulla and Nichol, Alex and Chu, Casey and Chen, Mark},
  title        = {Hierarchical Text{-}Conditional Image Generation with CLIP Latents},
  year         = {2022},
  eprint       = {2204.06125},
  archivePrefix= {arXiv},
  primaryClass = {cs.CV}
}

@inproceedings{reiser2021kilonerf,
  author    = {Reiser, Christian and Peng, Songyou and Liao, Yiyi and Geiger, Andreas},
  title     = {KiloNeRF: Speeding Up Neural Radiance Fields with Thousands of Tiny MLPs},
  booktitle = {Proceedings of the IEEE/CVF International Conference on Computer Vision (ICCV)},
  year      = {2021},
  pages     = {14335--14345}
}

@article{reiser2023merf,
  author  = {Reiser, Christian and Szeliski, Rick and Verbin, Dor and Srinivasan, Pratul and Mildenhall, Ben and Geiger, Andreas and Barron, Jon and Hedman, Peter},
  title   = {MERF: Memory{-}Efficient Radiance Fields for Real{-}Time View Synthesis in Unbounded Scenes},
  journal = {ACM Transactions on Graphics},
  volume  = {42},
  number  = {4},
  pages   = {1--12},
  year    = {2023}
}

@inproceedings{rombach2021geometryfree,
  author    = {Rombach, Robin and Esser, Patrick and Ommer, Bj{\"o}rn},
  title     = {Geometry{-}Free View Synthesis: Transformers and No 3D Priors},
  booktitle = {Proceedings of the IEEE/CVF International Conference on Computer Vision (ICCV)},
  year      = {2021},
  pages     = {14356--14366}
}

@inproceedings{saharia2022palette,
  author    = {Saharia, Chitwan and Chan, William and Chang, Huiwen and Lee, Chris and Ho, Jonathan and Salimans, Tim and Fleet, David and Norouzi, Mohammad},
  title     = {Palette: Image{-}to{-}Image Diffusion Models},
  booktitle = {Proceedings of SIGGRAPH},
  year      = {2022}
}

@article{saharia2022imagesr,
  author  = {Saharia, Chitwan and Ho, Jonathan and Chan, William and Salimans, Tim and Fleet, David J. and Norouzi, Mohammad},
  title   = {Image Super{-}Resolution via Iterative Refinement},
  journal = {IEEE Transactions on Pattern Analysis and Machine Intelligence},
  volume  = {45},
  number  = {4},
  pages   = {4713--4726},
  year    = {2022}
}

@inproceedings{sajjadi2022srt,
  author    = {Sajjadi, Mehdi SM. and Meyer, Henning and Pot, Etienne and Bergmann, Urs and Greff, Klaus and Radwan, Noha and Vora, Suhani and Lu{\v{c}}i{\'c}, Mario and Duckworth, Daniel and Dosovitskiy, Alexey and others},
  title     = {Scene Representation Transformer: Geometry{-}Free Novel View Synthesis through Set{-}Latent Scene Representations},
  booktitle = {Proceedings of the IEEE/CVF Conference on Computer Vision and Pattern Recognition (CVPR)},
  year      = {2022},
  pages     = {6229--6238}
}

@misc{shi2023zero123plus,
  author       = {Shi, Ruoxi and Chen, Hansheng and Zhang, Zhuoyang and Liu, Minghua and Xu, Chao and Wei, Xinyue and Chen, Linghao and Zeng, Chong and Su, Hao},
  title        = {Zero123++: A Single Image to Consistent Multi{-}View Diffusion Base Model},
  year         = {2023},
  eprint       = {2310.15110},
  archivePrefix= {arXiv},
  primaryClass = {cs.CV}
}

@misc{shi2023mvdream,
  author       = {Shi, Yichun and Wang, Peng and Ye, Jianglong and Mai, Long and Li, Kejie and Yang, Xiao},
  title        = {{MVDream}: Multi{-}View Diffusion for 3D Generation},
  year         = {2023},
  eprint       = {2308.16512},
  archivePrefix= {arXiv},
  primaryClass = {cs.CV}
}

@inproceedings{sinha2009piecewise,
  author    = {Sinha, Sudipta and Steedly, Drew and Szeliski, Rick},
  title     = {Piecewise Planar Stereo for Image{-}Based Rendering},
  booktitle = {Proceedings of the IEEE International Conference on Computer Vision (ICCV)},
  year      = {2009},
  pages     = {1881--1888}
}

@inproceedings{sitzmann2021lfn,
  author    = {Sitzmann, Vincent and Rezchikov, Semon and Freeman, Bill and Tenenbaum, Josh and Durand, Fredo},
  title     = {Light Field Networks: Neural Scene Representations with Single{-}Evaluation Rendering},
  booktitle = {Advances in Neural Information Processing Systems (NeurIPS)},
  volume    = {34},
  pages     = {19313--19325},
  year      = {2021}
}

@inproceedings{suhail2022gpnr,
  author    = {Suhail, Mohammed and Esteves, Carlos and Sigal, Leonid and Makadia, Ameesh},
  title     = {Generalizable Patch{-}Based Neural Rendering},
  booktitle = {European Conference on Computer Vision (ECCV)},
  year      = {2022},
  pages     = {156--174}
}

@inproceedings{sun2022dvgo,
  author    = {Sun, Cheng and Sun, Min and Chen, Hwann{-}Tzong},
  title     = {Direct Voxel Grid Optimization: Super{-}Fast Convergence for Radiance Fields Reconstruction},
  booktitle = {Proceedings of the IEEE/CVF Conference on Computer Vision and Pattern Recognition (CVPR)},
  year      = {2022},
  pages     = {5459--5469}
}

@article{elfwing2017sigmoid,
  title={Sigmoid-Weighted linear units for neural network function approximation in reinforcement learning. arXiv e-prints, art},
  author={Elfwing, S and Uchibe, E and Doya, K},
  journal={arXiv preprint arXiv:1702.03118},
  year={2017}
}

@inproceedings{rombach2022high,
  title={High-resolution image synthesis with latent diffusion models},
  author={Rombach, Robin and Blattmann, Andreas and Lorenz, Dominik and Esser, Patrick and Ommer, Bj{\"o}rn},
  booktitle={Proceedings of the IEEE/CVF conference on computer vision and pattern recognition},
  pages={10684--10695},
  year={2022}
}

@misc{szymanowicz2024flash3d,
  author       = {Szymanowicz, Stanislaw and Insafutdinov, Eldar and Zheng, Chuanxia and Campbell, Dylan and Henriques, Joao and Rupprecht, Christian and Vedaldi, Andrea},
  title        = {Flash3D: Feed{-}Forward Generalisable 3D Scene Reconstruction from a Single Image},
  year         = {2024},
  eprint       = {2403.00000},
  archivePrefix= {arXiv},
  primaryClass = {cs.CV}
}

@inproceedings{tang2024lgm,
  title={Lgm: Large multi-view gaussian model for high-resolution 3d content creation},
  author={Tang, Jiaxiang and Chen, Zhaoxi and Chen, Xiaokang and Wang, Tengfei and Zeng, Gang and Liu, Ziwei},
  booktitle={European Conference on Computer Vision},
  pages={1--18},
  year={2024},
  organization={Springer}
}

@article{vaswani2017attention,
  title={Attention is all you need},
  author={Vaswani, Ashish and Shazeer, Noam and Parmar, Niki and Uszkoreit, Jakob and Jones, Llion and Gomez, Aidan N and Kaiser, {\L}ukasz and Polosukhin, Illia},
  journal={Advances in neural information processing systems},
  volume={30},
  year={2017}
}

@inproceedings{tung2024megascenes,
  author    = {Tung, Joseph and Chou, Gene and Cai, Ruojin and Yang, Guandao and Zhang, Kai and Wetzstein, Gordon and Hariharan, Bharath and Snavely, Noah},
  title     = {MegaScenes: Scene{-}Level View Synthesis at Scale},
  booktitle = {Proceedings of the European Conference on Computer Vision (ECCV)},
  year      = {2024}
}

@inproceedings{verbin2022refnerf,
  author    = {Verbin, Dor and Hedman, Peter and Mildenhall, Ben and Zickler, Todd and Barron, Jonathan T. and Srinivasan, Pratul P.},
  title     = {{Ref{-}NeRF}: Structured View{-}Dependent Appearance for Neural Radiance Fields},
  booktitle = {Proceedings of the IEEE/CVF Conference on Computer Vision and Pattern Recognition (CVPR)},
  year      = {2022}
}

@inproceedings{voleti2025sv3d,
  author    = {Voleti, Vikram and Yao, Chun{-}Han and Boss, Mark and Letts, Adam and Pankratz, David and Tochilkin, Dmitry and Laforte, Christian and Rombach, Robin and Jampani, Varun},
  title     = {{SV3D}: Novel Multi{-}View Synthesis and 3D Generation from a Single Image Using Latent Video Diffusion},
  booktitle = {Proceedings of the European Conference on Computer Vision (ECCV)},
  year      = {2025},
  pages     = {439--457}
}

@misc{wang2023pflrm,
  author       = {Wang, Peng and Tan, Hao and Bi, Sai and Xu, Yinghao and Luan, Fujun and Sunkavalli, Kalyan and Wang, Wenping and Xu, Zexiang and Zhang, Kai},
  title        = {{PF{-}LRM}: Pose{-}Free Large Reconstruction Model for Joint Pose and Shape Prediction},
  year         = {2023},
  eprint       = {2311.12024},
  archivePrefix= {arXiv},
  primaryClass = {cs.CV}
}

@inproceedings{wang2024motionctrl,
  title={Motionctrl: A unified and flexible motion controller for video generation},
  author={Wang, Zhouxia and Yuan, Ziyang and Wang, Xintao and Li, Yaowei and Chen, Tianshui and Xia, Menghan and Luo, Ping and Shan, Ying},
  booktitle={ACM SIGGRAPH 2024 Conference Papers},
  pages={1--11},
  year={2024}
}

@misc{wang2021nerfmm,
  author       = {Wang, Zirui and Wu, Shangzhe and Xie, Weidi and Chen, Min and Prisacariu, Victor Adrian},
  title        = {NeRF{-}{-}: Neural Radiance Fields without Known Camera Parameters},
  year         = {2021},
  eprint       = {2102.07064},
  archivePrefix= {arXiv},
  primaryClass = {cs.CV}
}

@misc{watson2022novel,
  author       = {Watson, Daniel and Chan, William and Martin{-}Brualla, Ricardo and Ho, Jonathan and Tagliasacchi, Andrea and Norouzi, Mohammad},
  title        = {Novel View Synthesis with Diffusion Models},
  year         = {2022},
  eprint       = {2210.04628},
  archivePrefix= {arXiv},
  primaryClass = {cs.CV}
}

@misc{wei2024meshlrm,
  author       = {Wei, Xinyue and Zhang, Kai and Bi, Sai and Tan, Hao and Luan, Fujun and Deschaintre, Valentin and Sunkavalli, Kalyan and Su, Hao and Xu, Zexiang},
  title        = {MeshLRM: Large Reconstruction Model for High{-}Quality Mesh},
  year         = {2024},
  eprint       = {2404.12385},
  archivePrefix= {arXiv},
  primaryClass = {cs.CV}
}

@misc{xie2024lrmzero,
  author       = {Xie, Desai and Bi, Sai and Shu, Zhixin and Zhang, Kai and Xu, Zexiang and Zhou, Yi and Pirk, S{\"o}ren and Kaufman, Arie and Sun, Xin and Tan, Hao},
  title        = {LRM{-}Zero: Training Large Reconstruction Models with Synthesized Data},
  year         = {2024},
  eprint       = {2406.09371},
  archivePrefix= {arXiv},
  primaryClass = {cs.CV}
}

@inproceedings{xu2022pointnerf,
  author    = {Xu, Qiangeng and Xu, Zexiang and Philip, Julien and Bi, Sai and Shu, Zhixin and Sunkavalli, Kalyan and Neumann, Ulrich},
  title     = {Point{-}NeRF: Point{-}Based Neural Radiance Fields},
  booktitle = {Proceedings of the IEEE/CVF Conference on Computer Vision and Pattern Recognition (CVPR)},
  year      = {2022},
  pages     = {5438--5448}
}

@misc{xu2023dmv3d,
  author       = {Xu, Yinghao and Tan, Hao and Luan, Fujun and Bi, Sai and Wang, Peng and Li, Jiahao and Shi, Zifan and Sunkavalli, Kalyan and Wetzstein, Gordon and Xu, Zexiang and Zhang, Kai},
  title        = {DMV3D: Denoising Multi{-}View Diffusion Using 3D Large Reconstruction Model},
  year         = {2023},
  eprint       = {2311.00000},
  archivePrefix= {arXiv},
  primaryClass = {cs.CV}
}

@misc{yan2024streetcrafterstreetviewsynthesis,
  author       = {Yan, Yunzhi and Xu, Zhen and Lin, Haotong and Jin, Haian and Guo, Haoyu and Wang, Yida and Zhan, Kun and Lang, Xianpeng and Bao, Hujun and Zhou, Xiaowei and Peng, Sida},
  title        = {StreetCrafter: Street View Synthesis with Controllable Video Diffusion Models},
  year         = {2024},
  eprint       = {2412.13188},
  archivePrefix= {arXiv},
  primaryClass = {cs.CV}
}

@misc{yang2023consistnet,
  author       = {Yang, Jiayu and Cheng, Ziang and Duan, Yunfei and Ji, Pan and Li, Hongdong},
  title        = {ConsistNet: Enforcing 3D Consistency for Multi{-}View Images Diffusion},
  year         = {2023},
  eprint       = {2310.00000},
  archivePrefix= {arXiv},
  primaryClass = {cs.CV}
}

@misc{ye2023consistent,
  author       = {Ye, Jianglong and Wang, Peng and Li, Kejie and Shi, Yichun and Wang, Heng},
  title        = {Consistent{-}1{-}to{-}3: Consistent Image to 3D View Synthesis via Geometry{-}Aware Diffusion Models},
  year         = {2023},
  eprint       = {2310.03020},
  archivePrefix= {arXiv},
  primaryClass = {cs.CV}
}

@misc{yu2024viewcrafter,
  author       = {Yu, Wangbo and Xing, Jinbo and Yuan, Li and Hu, Wenbo and Li, Xiaoyu and Huang, Zhipeng and Gao, Xiangjun and Wong, Tien{-}Tsin and Shan, Ying and Tian, Yonghong},
  title        = {ViewCrafter: Taming Video Diffusion Models for High{-}Fidelity Novel View Synthesis},
  year         = {2024},
  eprint       = {2409.02048},
  archivePrefix= {arXiv},
  primaryClass = {cs.CV}
}

@inproceedings{zhang2022differentiablepoint,
  author    = {Zhang, Qiang and Baek, Seung{-}Hwan and Rusinkiewicz, Szymon and Heide, Felix},
  title     = {Differentiable Point{-}Based Radiance Fields for Efficient View Synthesis},
  booktitle = {SIGGRAPH Asia Conference Papers},
  year      = {2022},
  pages     = {1--12}
}

@inproceedings{zheng2024free3d,
  author    = {Zheng, Chuanxia and Vedaldi, Andrea},
  title     = {Free3D: Consistent Novel View Synthesis without 3D Representation},
  booktitle = {Proceedings of the IEEE/CVF Conference on Computer Vision and Pattern Recognition (CVPR)},
  year      = {2024},
  pages     = {9720--9731}
}

@inproceedings{zhou2018stereomagnification,
  author    = {Zhou, Tinghui and Tucker, Richard and Flynn, John and Fyffe, Graham and Snavely, Noah},
  title     = {Stereo Magnification: Learning View Synthesis Using Multiplane Images},
  booktitle = {SIGGRAPH},
  year      = {2018}
}

@inproceedings{mildenhall2020nerf,
  author    = {Ben Mildenhall and Pratul P. Srinivasan and Matthew Tancik and Jonathan T. Barron and Ravi Ramamoorthi and Ren Ng},
  title     = {{N}e{RF}: Representing Scenes as Neural Radiance Fields for View Synthesis},
  booktitle = {European Conference on Computer Vision},
  pages     = {405--421},
  year      = {2020}
}

@misc{poole2022dreamfusion,
  author    = {Ben Poole and Ajay Jain and Jonathan T. Barron and Ben Mildenhall},
  title     = {{D}reamfusion: Text-to-{3D} using {2D} Diffusion},
  howpublished = {arXiv},
  year      = {2022}
}

@inproceedings{reconfusion,
  author    = {Rundi Wu and Ben Mildenhall and Philipp Henzler and Keunhong Park and Ruiqi Gao and Daniel Watson and Pratul P. Srinivasan and Dor Verbin and Jonathan T. Barron and Ben Poole and others},
  title     = {{R}econfusion: {3D} Reconstruction with Diffusion Priors},
  booktitle = {Proceedings of the IEEE/CVF Conference on Computer Vision and Pattern Recognition},
  pages     = {21551--21561},
  year      = {2024}
}

@misc{jin2024lvsmlargeviewsynthesis,
  author    = {Haian Jin and Hanwen Jiang and Hao Tan and Kai Zhang and Sai Bi and Tianyuan Zhang and Fujun Luan and Noah Snavely and Zexiang Xu},
  title     = {{LVSM}: {A} Large View Synthesis Model with Minimal {3D} Inductive Bias},
  year      = {2024}
}

@inproceedings{4dim,
  title={Controlling space and time with diffusion models},
  author={Watson, Daniel and Saxena, Saurabh and Li, Lala and Tagliasacchi, Andrea and Fleet, David J},
  booktitle={The Thirteenth International Conference on Learning Representations},
  year={2024}
}

@misc{sargent2023zeronvs,
  author    = {Kyle Sargent and Zizhang Li and Tanmay Shah and Charles Herrmann and Hong-Xing Yu and Yunzhi Zhang and Eric Ryan Chan and Dmitry Lagun and Li Fei-Fei and Deqing Sun and others},
  title     = {{Z}ero{NVS}: Zero-Shot 360-Degree View Synthesis from a Single Real Image},
  howpublished = {arXiv preprint arXiv:2310.17994},
  year      = {2023}
}

@inproceedings{voleti2024sv3d,
  author    = {Vikram Voleti and Chun-Han Yao and Mark Boss and Adam Letts and David Pankratz and Dmitrii Tochilkin and Christian Laforte and Robin Rombach and Varun Jampani},
  title     = {{SV3D}: Novel Multi-View Synthesis and {3D} Generation from a Single Image using Latent Video Diffusion},
  booktitle = {European Conference on Computer Vision},
  year      = {2024}
}

@inproceedings{ho2020denoising,
  author    = {Jonathan Ho and Ajay Jain and Pieter Abbeel},
  title     = {Denoising Diffusion Probabilistic Models},
  booktitle = {Advances in Neural Information Processing Systems},
  volume    = {33},
  pages     = {6840--6851},
  year      = {2020}
}

@article{blattmann2023stable,
  title={Stable video diffusion: Scaling latent video diffusion models to large datasets},
  author={Blattmann, Andreas and Dockhorn, Tim and Kulal, Sumith and Mendelevitch, Daniel and Kilian, Maciej and Lorenz, Dominik and Levi, Yam and English, Zion and Voleti, Vikram and Letts, Adam and others},
  journal={arXiv preprint arXiv:2311.15127},
  year={2023}
}

@inproceedings{suhail2022light,
  author    = {Mohammed Suhail and Carlos Esteves and Leonid Sigal and Ameesh Makadia},
  title     = {Light Field Neural Rendering},
  booktitle = {Proceedings of the IEEE/CVF Conference on Computer Vision and Pattern Recognition},
  pages     = {8269--8279},
  year      = {2022}
}

@inproceedings{wan2018confnet,
  title={Confnet: predict with confidence},
  author={Wan, Sheng and Wu, Tung-Yu and Wong, Wing H and Lee, Chen-Yi},
  booktitle={2018 IEEE International Conference on Acoustics, Speech and Signal Processing (ICASSP)},
  pages={2921--2925},
  year={2018},
  organization={IEEE}
}

@inproceedings{peebles2023scalable,
  title={Scalable diffusion models with transformers},
  author={Peebles, William and Xie, Saining},
  booktitle={Proceedings of the IEEE/CVF international conference on computer vision},
  pages={4195--4205},
  year={2023}
}

@inproceedings{gregor2014deep,
  author    = {Karol Gregor and Ivo Danihelka and Andriy Mnih and Charles Blundell and Daan Wierstra},
  title     = {Deep Autoregressive Networks},
  booktitle = {International Conference on Machine Learning},
  pages     = {1242--1250},
  year      = {2014}
}

@inproceedings{van2016pixel,
  author    = {A{\"a}ron Van Den Oord and Nal Kalchbrenner and Koray Kavukcuoglu},
  title     = {Pixel Recurrent Neural Networks},
  booktitle = {International Conference on Machine Learning},
  pages     = {1747--1756},
  year      = {2016}
}

@article{van2016conditional,
  title={Conditional image generation with pixelcnn decoders},
  author={Van den Oord, Aaron and Kalchbrenner, Nal and Espeholt, Lasse and Vinyals, Oriol and Graves, Alex and others},
  journal={Advances in neural information processing systems},
  volume={29},
  year={2016}
}

@article{deng2024autoregressive,
  author    = {Haoge Deng and Ting Pan and Haiwen Diao and Zhengxiong Luo and Yufeng Cui and Huchuan Lu and Shiguang Shan and Yonggang Qi and Xinlong Wang},
  title     = {Autoregressive Video Generation without Vector Quantization},
  journal   = {arXiv preprint arXiv:2412.14169},
  year      = {2024}
}

@inproceedings{li2023mage,
  title={Mage: Masked generative encoder to unify representation learning and image synthesis},
  author={Li, Tianhong and Chang, Huiwen and Mishra, Shlok and Zhang, Han and Katabi, Dina and Krishnan, Dilip},
  booktitle={Proceedings of the IEEE/CVF Conference on Computer Vision and Pattern Recognition},
  pages={2142--2152},
  year={2023}
}

@inproceedings{chang2022maskgit,
  title={Maskgit: Masked generative image transformer},
  author={Chang, Huiwen and Zhang, Han and Jiang, Lu and Liu, Ce and Freeman, William T},
  booktitle={Proceedings of the IEEE/CVF conference on computer vision and pattern recognition},
  pages={11315--11325},
  year={2022}
}

@article{li2024autoregressive,
  author    = {Tianhong Li and Yonglong Tian and He Li and Mingyang Deng and Kaiming He},
  title     = {Autoregressive Image Generation without Vector Quantization},
  journal   = {Advances in Neural Information Processing Systems},
  volume    = {37},
  pages     = {56424--56445},
  year      = {2024}
}

@inproceedings{wang2021ibrnet,
  title={Ibrnet: Learning multi-view image-based rendering},
  author={Wang, Qianqian and Wang, Zhicheng and Genova, Kyle and Srinivasan, Pratul P and Zhou, Howard and Barron, Jonathan T and Martin-Brualla, Ricardo and Snavely, Noah and Funkhouser, Thomas},
  booktitle={Proceedings of the IEEE/CVF conference on computer vision and pattern recognition},
  pages={4690--4699},
  year={2021}
}

@inproceedings{yu2020pixelnerf,
  author    = {Alex Yu and Vickie Ye and Matthew Tancik and Angjoo Kanazawa},
  title     = {{pixelNeRF}: Neural radiance fields from one or few images},
  booktitle = {CVPR},
  year      = {2021}
}

@inproceedings{chen2021mvsnerf,
  title={Mvsnerf: Fast generalizable radiance field reconstruction from multi-view stereo},
  author={Chen, Anpei and Xu, Zexiang and Zhao, Fuqiang and Zhang, Xiaoshuai and Xiang, Fanbo and Yu, Jingyi and Su, Hao},
  booktitle={Proceedings of the IEEE/CVF international conference on computer vision},
  pages={14124--14133},
  year={2021}
}

@article{zhou2025stable,
  author    = {Jensen Jinghao Zhou and Hang Gao and Vikram Voleti and Aaryaman Vasishta and Chun-Han Yao and Mark Boss and Philip Torr and Christian Rupprecht and Varun Jampani},
  title     = {Stable Virtual Camera: Generative View Synthesis with Diffusion Models},
  journal   = {arXiv preprint arXiv:2503.14489},
  year      = {2025}
}

@misc{mercier2024hexagen3d,
  author    = {Antoine Mercier and Ramin Nakhli and Mahesh Reddy and Rajeev Yasarla and Hong Cai and Fatih Porikli and Guillaume Berger},
  title     = {{H}exa{G}en{3D}: {S}table{D}iffusion is Just One Step Away from Fast and Diverse Text-to-{3D} Generation},
  howpublished = {arXiv preprint arXiv:2401.07727},
  year      = {2024}
}

@misc{sd3,
  author       = {{Stability AI}},
  title        = {Stable Diffusion 3 (technical preview)},
  howpublished = {Press release},
  year         = {2024}
}

@article{ye2024no,
  title={No pose, no problem: Surprisingly simple 3d gaussian splats from sparse unposed images},
  author={Ye, Botao and Liu, Sifei and Xu, Haofei and Li, Xueting and Pollefeys, Marc and Yang, Ming-Hsuan and Peng, Songyou},
  journal={arXiv preprint arXiv:2410.24207},
  year={2024}
}

@inproceedings{ling2024dl3dv,
  title={Dl3dv-10k: A large-scale scene dataset for deep learning-based 3d vision},
  author={Ling, Lu and Sheng, Yichen and Tu, Zhi and Zhao, Wentian and Xin, Cheng and Wan, Kun and Yu, Lantao and Guo, Qianyu and Yu, Zixun and Lu, Yawen and others},
  booktitle={Proceedings of the IEEE/CVF Conference on Computer Vision and Pattern Recognition},
  pages={22160--22169},
  year={2024}
}

@article{loshchilov2017decoupled,
  title={Decoupled weight decay regularization},
  author={Loshchilov, Ilya and Hutter, Frank},
  journal={arXiv preprint arXiv:1711.05101},
  year={2017}
}

@article{xu2024depthsplat,
  title={Depthsplat: Connecting gaussian splatting and depth},
  author={Xu, Haofei and Peng, Songyou and Wang, Fangjinhua and Blum, Hermann and Barath, Daniel and Geiger, Andreas and Pollefeys, Marc},
  journal={arXiv preprint arXiv:2410.13862},
  year={2024}
}

@misc{kingma2013auto,
  title={Auto-encoding variational bayes},
  author={Kingma, Diederik P and Welling, Max and others},
  year={2013},
  publisher={Banff, Canada}
}
\bibliographystyle{abbrv}
\clearpage

\appendix
\begin{center}
{\bf{\large{Supplement to ``UMAMI: Unifying Masked Autoregressive Models and Deterministic Rendering for View Synthesis''}}}
\end{center}

\section{Failure cases}

 \begin{figure}[!ht]
\centering
\includegraphics[width=\textwidth]{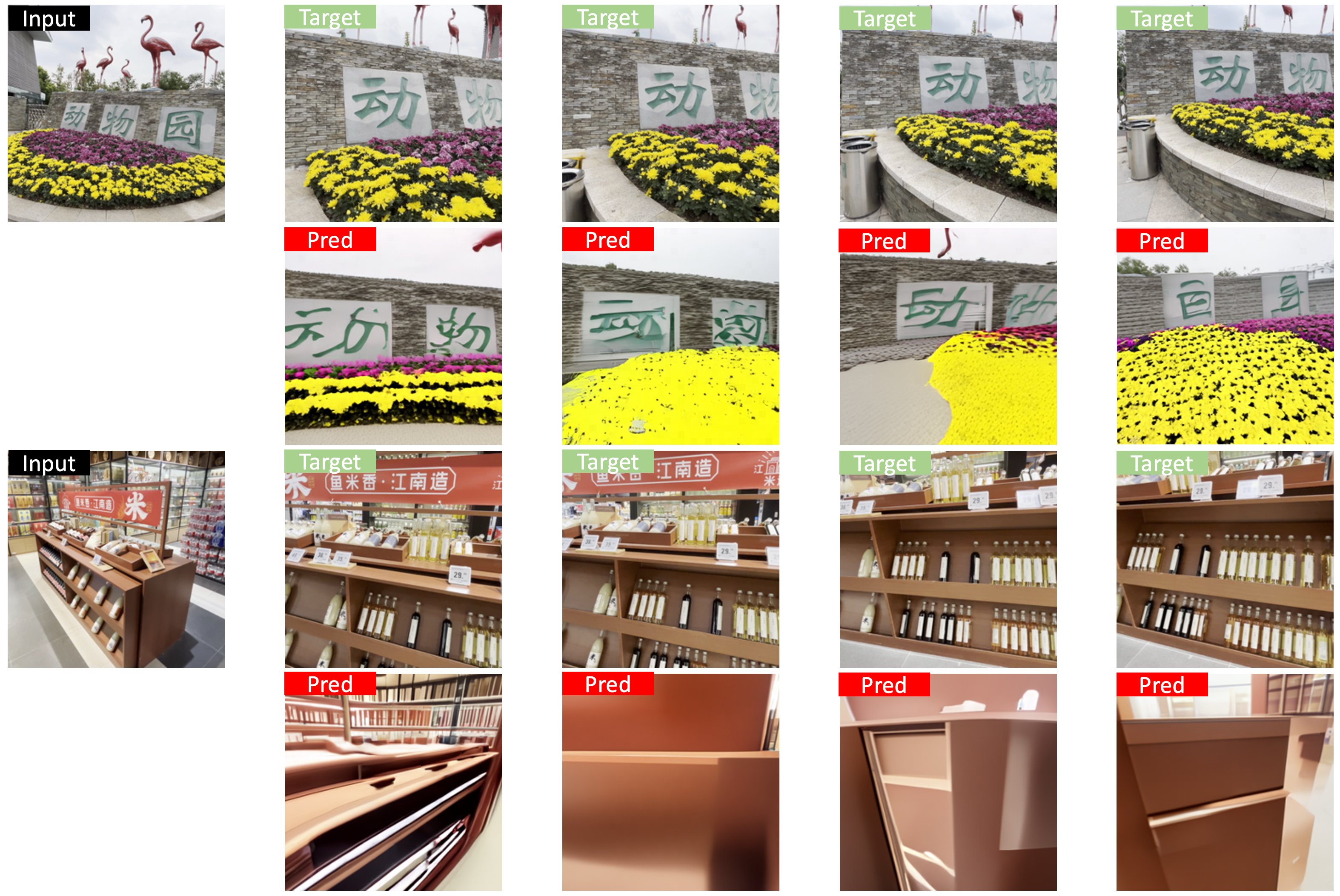}
\caption{\textbf{Failure cases.} Our method may produce noticeable artifacts when target camera poses are too distant from the input view. Increasing the scale of training data and model parameters could improve the robustness of \mname.}
\label{fig:supp-failure}

\end{figure}

\section{Related works}
\subsection{Feed-forward deterministic NVS methods}
Early generalizable methods for Novel View Synthesis (NVS) demonstrated the potential of neural networks, trained across various scenes, to enable fast inference of novel views or underlying 3D representations in a feed-forward manner. Prominent examples include PixelNeRF~\citep{yu2020pixelnerf}, MVSNeRF~\citep{chen2021mvsnerf}, and IBRNet~\citep{wang2021ibrnet}, which typically predict volumetric 3D representations by incorporating 3D-specific priors like epipolar geometry or plane sweep cost volumes. Subsequent research has extended these capabilities, improving performance particularly under challenging conditions such as sparse input views~\citep{liu2022neuralray, johari2022geonerfgeneralizingnerfgeometry,jiang2024forge,jiang2023leap}, and adapting these techniques for emerging representations like 3D Gaussian Splatting (3DGS)~\cite{charatan2024pixelsplat, szymanowicz2024flash3d,chen2024mvsplat,tang2024lgm}.

Recently, 3D Large Reconstruction Models (LRMs) have emerged~\citep{hong2024lrmlargereconstructionmodel,li2023instant3d, wang2023pflrm,xu2023dmv3d, wei2024meshlrm,xie2024lrmzero}, leveraging the power of scalable transformer architectures~\citep{vaswani2017attention} trained on extensive datasets to learn generic 3D priors. While these methods successfully avoid explicit architectural reliance on epipolar projection or cost volumes, they still typically depend on pre-defined 3D representations such as tri-plane NeRFs, meshes, or 3DGS, along with their corresponding rendering equations. This reliance can limit their flexibility and overall potential. 

An alternative line of work attempts to directly learn a geometry-free rendering function~\citep{suhail2022gpnr, sajjadi2022srt, sitzmann2021lfn, rombach2021geometryfree, kulhanek2022viewformer}. However, these approaches often face limitations in model capacity and scalability, which can hinder their ability to capture high-frequency details. Notably, Scene Representation Transformers (SRT)~\citep{sajjadi2022srt} aimed to avoid explicit, handcrafted 3D representations by learning a latent scene representation via a transformer, an objective shared by our encoder-decoder architecture. Despite this similarity, certain design choices in SRT, such as its CNN-based token extractor and the use of cross-attention in the decoder, have been shown to lead to less effective performance. To address the issue, LVSM~\cite{jin2024lvsmlargeviewsynthesis} proposes a method that is fully transformer-based, leveraging bidirectional self-attention for enhanced representational power. Furthermore, they introduce a novel and more scalable decoder-only architecture that directly learns the NVS function with minimal 3D inductive bias and without relying on an intermediate latent representation. 

Our proposed method adopts the versatile and scalable decoder-only transformer backbone from LVSM, which has demonstrated its efficacy in NVS tasks by leveraging a data-driven approach with minimal handcrafted 3D inductive bias. However, a crucial distinction lies in the nature of our approach: unlike the deterministic LVSM, our method is generative. We aim to address the inherent limitations of deterministic methods by harnessing the generative capabilities of masked autoregressive diffusion models in an efficient manner.  

\subsection{Generative-based NVS methods}
The pursuit of generative-based (NVS) has recently seen significant advancements through the integration of diffusion models, drawing inspiration from successes in broader NVS~\citep{sitzmann2021lfn,sajjadi2022srt} and generative image-to-image tasks~\citep{saharia2022palette,ramesh2022hierarchical,saharia2022imagesr}.

An early exploration in this domain was 3DiM~\citep{watson2022novel}, which trained image-to-image diffusion models for object-level multi-view rendering without explicit 3D representations. However, by training from scratch on limited 3D data, 3DiM's applicability was restricted to category-specific scenarios and lacked zero-shot generalization capabilities. Building on this, Zero-1-to-3~\citep{liu2023zero1to3} adopted a similar geometry-free pipeline but significantly improved generalization and output quality by fine-tuning a pretrained 2D diffusion model on a larger 3D object dataset~\citep{deitke2023objaverse}. Despite these improvements, a key challenge for Zero-1-to-3 and other early image-based diffusion models for NVS (e.g., for distant viewpoints~\citep{shi2023zero123plus}) was multi-view inconsistency, as they typically generated each target view independently and probabilistically, leading to jitter or inconsistencies when rendering a camera trajectory.

To address this multi-view inconsistency, subsequent research diverged into several directions. One line of work focused on integrating explicit 3D inductive biases—such as 3D representations or epipolar attention—into the diffusion denoising process. Examples include SyncDreamer~\citep{liu2023syncdreamer}, ConsistNet~\citep{yang2023consistnet}, Consistent-1-to-N~\citep{ye2023consistent}, and MegaScenes~\citep{tung2024megascenes}, though these often came at the cost of increased computation. Another set of approaches, including Instant3D~\citep{li2023instant3d}, MVDream~\citep{shi2023mvdream}, and Wonder3D~\citep{long2023wonder3d}, aimed to predict a single grid of multiple, specific views simultaneously. While this improved consistency across those fixed views, it sacrificed the ability for fine-grained camera control. Works like MVDream~\citep{shi2023mvdream}, SyncDreamer~\citep{liu2023syncdreamer}, and more recently HexGen3D~\citep{mercier2024hexagen3d}, generate multiple fixed views from a conditional image but do not support arbitrary viewpoint selection. To achieve consistent 3D object geometry from these image-based models, further steps like NeRF distillation, using techniques such as Score Distillation Sampling (SDS)~\citep{poole2022dreamfusion,sargent2023zeronvs} or direct optimization on sampled images~\citep{reconfusion,gao2024cat3d}, are often necessary. However, distillation techniques such as SDS can introduce substantial computational overhead due to test-time optimization.

More recently, a promising trend has emerged with models that jointly predict multiple target views while maintaining accurate camera control and ensuring view consistency, often through mechanisms like cross-view attention. This category includes methods such as Free3D~\citep{zheng2024free3d}, EscherNet~\citep{kong2024eschernet}, CAT3D~\citep{gao2024cat3d}, and SV3D~\citep{voleti2025sv3d}. Several video model-based approaches~\citep{wang2024motionctrl, he2024cameractrl, yu2024viewcrafter,yan2024streetcrafterstreetviewsynthesis, zhou2025stable} also fall into this paradigm, increasing NVS performance. Despite these advancements, achieving high-quality generation with these recent models often necessitates substantial computational resources and extensive training data. Furthermore, their reliance on full-image iterative sampling typically results in slow inference times, limiting practical applicability. Our proposed method, \mname, addresses this critical issue by enabling photorealistic novel view rendering while maintaining efficient inference times.

\section{Implementation details}
\subsection{Hyperparamters}
\begin{table}[h]
\centering
\caption{Hyperparameters for training \mname. We use the same set of hyperparameters for both RealEstate10K and DL3DV experiments.}
\begin{tabular}{lll}
\toprule
\textbf{Component} & \textbf{Parameter} & \textbf{Value} \\
\midrule
\multirow{3}{*}{Image Tokenizer} 
  & Image size & 256 \\
  & Patch size & 8 \\
  & Channels & 9 (3 RGB + 6 for Plücker) \\
\midrule
\multirow{5}{*}{Transformer}
  & Layers & 24 \\
  & Hidden dim & 768 \\
  & Head dim & 64 \\
  & QK Norm & True \\
\midrule
\multirow{9}{*}{Training}
  & Batch size / GPU & 4 \\
  & Num GPUS & 8 \\
  & Learning rate & 0.0002 \\
  & Optimizer ($\beta_1$, $\beta_2$) & (0.9, 0.95) \\
  & Grad clip norm & 3.0 \\
  & Mixed precision & True \\
  & Weight decay & 0.02 \\
  & Train steps & 100k \\
  & Warmup steps & 1000 \\
\midrule
\multirow{2}{*}{Data Setup}
  & Input / Target views & 1 to 2 / 1 to 3 \\
  & Center Crop & True \\
\midrule
\multirow{5}{*}{Loss Weights}
  & L2 loss & 1.0 \\
  & LPIPS loss & 0.0 \\
  & Perceptual loss & 0.5 \\
  & Diffusion loss & 10 \\
  & Confidence loss & 1 \\  
\bottomrule
\end{tabular}
\label{tab:hyperparams}
\end{table}
We report the hyperparameters used in Table \ref{tab:hyperparams}. 

\subsection{Algorithm}
We describe the sampling process of \mname in Algorithm \ref{alg:inference}.

\begin{algorithm}[H]
\caption{Hybrid Inference in UMAMI}
\label{alg:inference}
\begin{algorithmic}[1]
\Require Trained model, context views $\{(I_{\text{ctx}}, \pi_{\text{ctx}})\}$, target pose $\pi_{\text{tgt}}$, threshold $\tau$, max unmasking steps $T_{\max}$
\State Tokenize context views into $\mathbf{c}$, initialize target tokens $\mathbf{x}$ with masked tokens
\State Encode $(\mathbf{c}, \mathbf{x})$ with Transformer to obtain latent $\mathbf{z}$
\State Predict confidence map $\mathbf{s}_p$ and patch-level scores $\mathbf{s}$
\State Partition target tokens:
    \begin{itemize}
        \item Deterministic tokens: $\mathbf{x}_D \leftarrow \{x_i \mid s_i \geq \tau\}$
        \item Stochastic tokens: $\mathbf{x}_S \leftarrow \{x_i \mid s_i < \tau\}$
    \end{itemize}
\State Predict $\mathbf{x}_D$ in one pass using deterministic head: $\hat{\mathbf{x}}_D = \phi(\mathbf{z}_D)$
\State Compute sampling steps: $T_S = \lceil |x_S| / |x| \cdot T_{\max} \rceil$
\For{$t = T_S$ \textbf{to} $1$}
    \State Sample random unmasked set $\mathbf{x}_t \subset \mathbf{x}_S$ following a cosine scheduler.
    \State Update $\mathbf{x}_t$ by DDPM sampling using $\varphi$ head.
\EndFor
\State Merge $\hat{\mathbf{x}}_D$ and $\hat{\mathbf{x}}_S$ into full target image $\hat{I}_{\text{tgt}}$
\end{algorithmic}
\end{algorithm}

\section{Additional quantitative results}
\subsection{Multiple images generation}
\begin{table}[h]
\centering
\caption{Multi-view generation results on RealEstate10K.}
\begin{tabular}{lcccc}
\toprule
\textbf{Dataset} & \textbf{\# gen views} & \textbf{PSNR} $\uparrow$ & \textbf{LPIPS} $\downarrow$ & \textbf{SSIM} $\uparrow$ \\
\midrule
\multirow{2}{*}{Re10K-2views-Extra} 
& 1 & 28.95 & 0.107 & 0.897 \\
& 3 & 28.65 & 0.109 & 0.892 \\
\midrule
\multirow{2}{*}{Re10K-2views-Interp} 
& 1 & 28.85 & 0.101 & 0.899 \\
& 3 & 28.52 & 0.105 & 0.894 \\
\bottomrule
\end{tabular}
\label{tab:multigen}
\end{table}
As shown in Table~\ref{tab:hyperparams}, our model is trained to predict up to three target views simultaneously. This joint prediction encourages consistency across generated images. In Table~\ref{tab:multigen}, we report results for generating one and three views. The generation quality is comparable across both settings. Notably, we use a fixed number of unmasking steps ($T_{\max} = 32$) for all cases, which means generating multiple views in parallel can improve inference efficiency without sacrificing quality.

% \input{tables/overlap}
% \subsection{Varying overlap ratios}
% Following NoPoSplat~\citep{ye2024no}, we evaluate \mname on Re10k test set with varying camera overlaps based on ratio of image overlap: small ($0.05\%$-$0.3\%$), medium ($0.3\%$ - $0.55\%$), and large ($0.55\%$ - $0.8\%$), determined using dense feature matching method, RoMA~\citep{edstedt2024roma}. The results, shown in the Table~\ref{tab:overlap}, demonstrate that \mname outperforms NoPoSplat on all metrics and validation sets. As with NoPoSplat, we also observed that performance improves as the overlap ratio increases, which suggests less scene occlusion. Notably, our method remains robust even with a small camera overlap, outperforming NoPoSplat across all metrics. This confirms \mname's effectiveness across various datasets and overlap ratios.

\section{Additional qualitative results}

\begin{figure}[!ht]
\centering
\includegraphics[width=\textwidth]{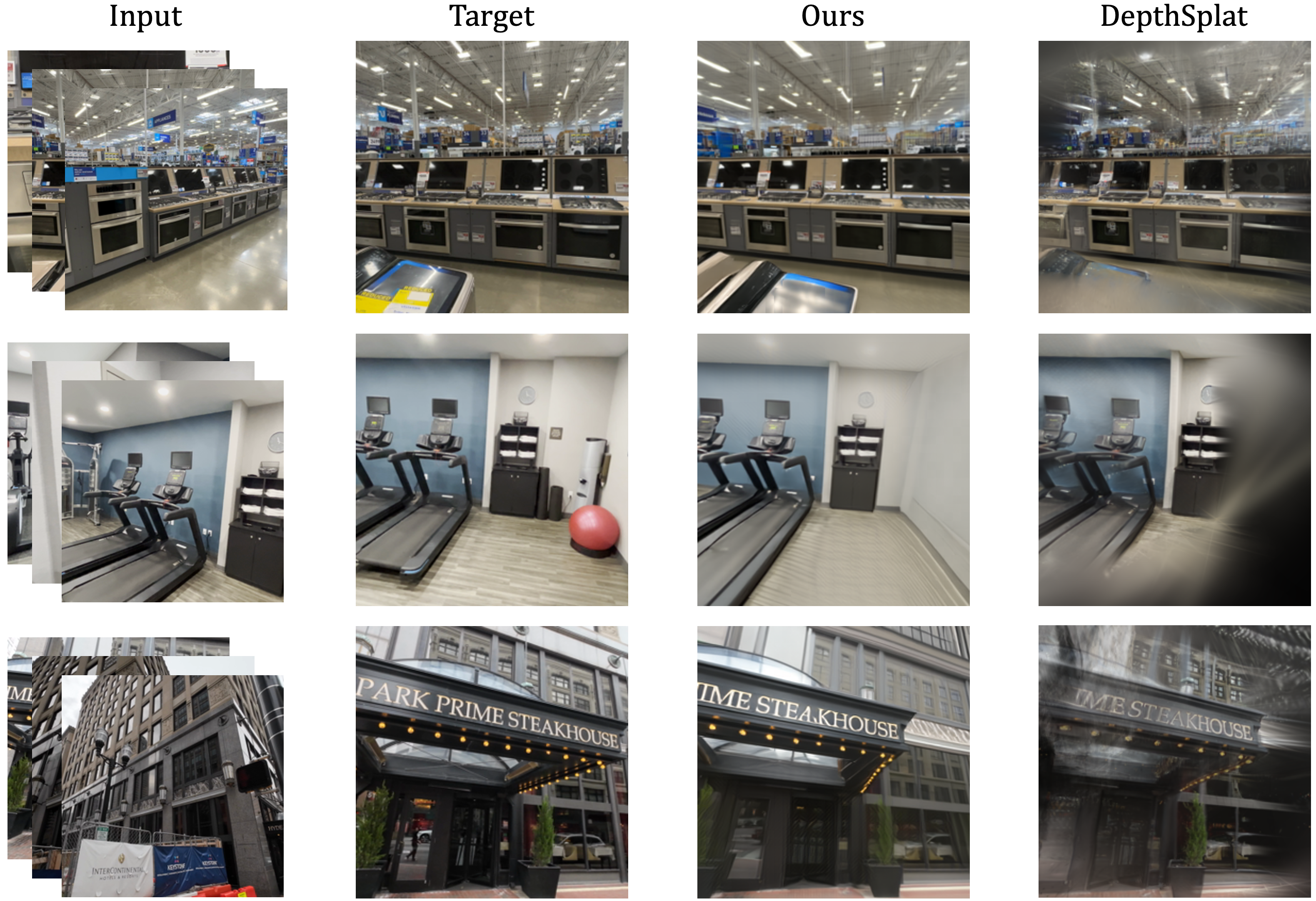}
\caption{Additional qualitative comparisons on DL3DV dataset.}
\label{fig:supp-dl3dv}

\end{figure}

\begin{figure}[!ht]
\centering
\includegraphics[width=\textwidth]{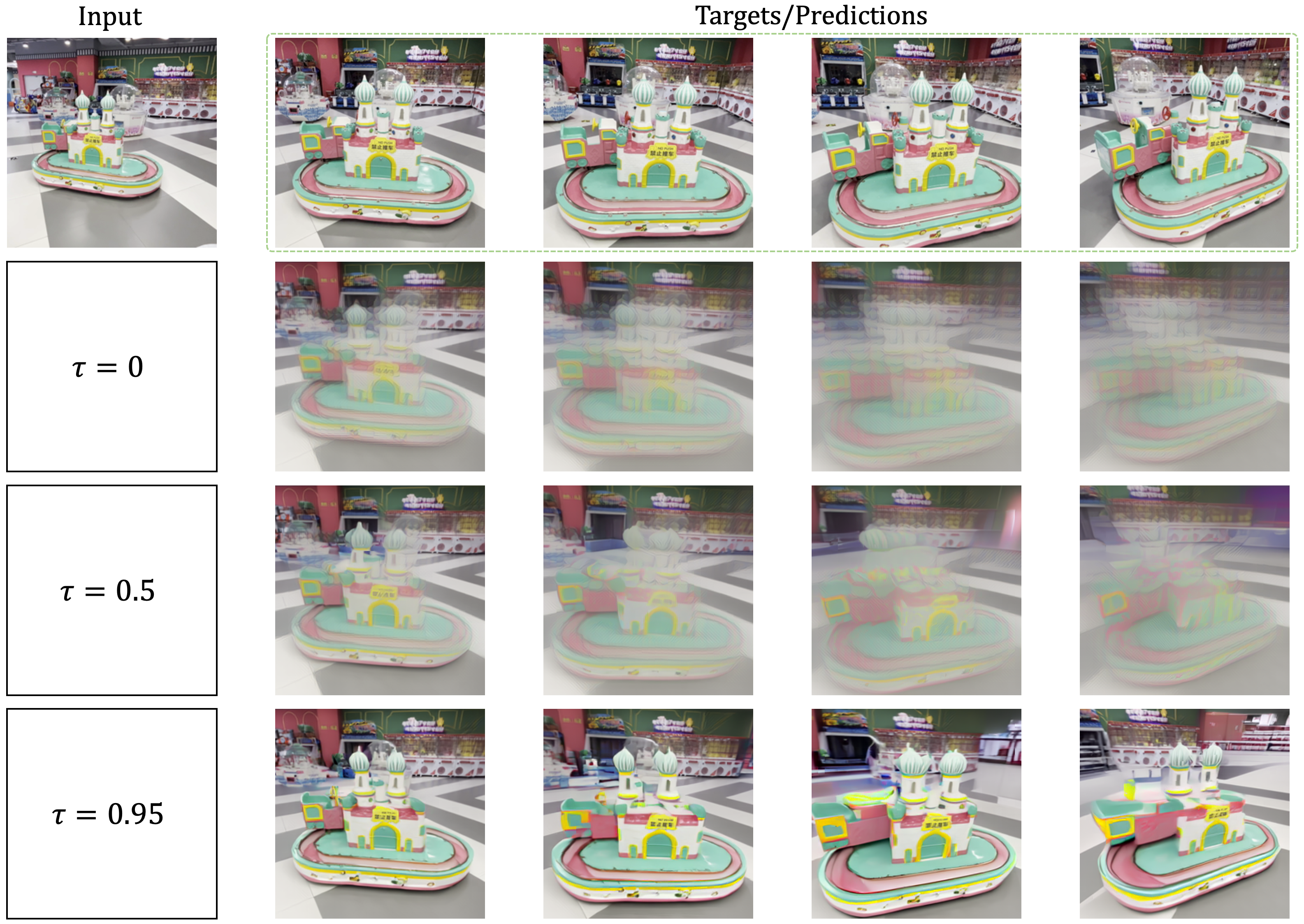}
\caption{Impact of the diffusion threshold hyperparameter $\tau$ on rendering outcomes. The top row shows the single input view alongside four corresponding target views. The subsequent rows (2-4) illustrate the results as $\tau$ is incrementally increased. While a lower $\tau$ promotes deterministic behavior and faster inference, higher values of $\tau$ lead to notably sharper image rendering quality.}
\label{fig:supp-thresholding}

\end{figure}

Additional qualitative evaluations are presented on the DL3DV dataset~\citep{ling2024dl3dv}, where our method is compared against DepthSplat~\citep{xu2024depthsplat} under a three-view input configuration (Figure~\ref{fig:supp-dl3dv}). As depicted \mname demonstrates notably sharp rendering, particularly in unobserved regions. This is achieved by leveraging its generative capabilities to synthesize plausible details unobserved region of input images.

Furthermore, to investigate the impact of the diffusion threshold hyperparameter, $\tau$, on \mname's performance, its value was systematically varied, with findings illustrated in Figure~\ref{fig:supp-thresholding}. An initial setting of $\tau=0$, corresponding to a fully deterministic operation of \mname, achieved rapid inference. However, this configuration resulted in image blurring, an artifact attributable to unobserved regions in the input view. Progressively increasing $\tau$ to $0.5$ and subsequently to $0.95$ yielded a significant enhancement in rendering quality. This improvement, however, was accompanied by an increase in running time. Finally, to demonstrate the complete sampling dynamics of our method, the unmasking processes for $\tau=0.95$ and for full unmasking diffusion process ($\tau=1$) are presented in the supplementary video.
\clearpage

%%%%%%%%%%%%%%%%%%%%%%%%%%%%%%%%%%%%%%%%%%%%%%%%%%%%%%%%%%%%

\newpage
\section*{NeurIPS Paper Checklist}

\begin{enumerate}

\item {\bf Claims}
    \item[] Question: Do the main claims made in the abstract and introduction accurately reflect the paper's contributions and scope?
    \item[] Answer: \answerYes{} % Replace by \answerYes{}, \answerNo{}, or \answerNA{}.
    \item[] Justification: In the abstract, we mainly claimed that our method can achieve state-of-the-art accuracy and can achieve faster speed compared to generative methods. There are supported with experimental evidence in our main paper.

\item {\bf Limitations}
    \item[] Question: Does the paper discuss the limitations of the work performed by the authors?
    \item[] Answer: \answerYes{} % Replace by \answerYes{}, \answerNo{}, or \answerNA{}.
    \item[] Justification: We have provided a dedicated subsection that discusses the limitations of our approach.

\item {\bf Theory assumptions and proofs}
    \item[] Question: For each theoretical result, does the paper provide the full set of assumptions and a complete (and correct) proof?
    \item[] Answer: \answerNA{} % Replace by \answerYes{}, \answerNo{}, or \answerNA{}.
    \item[] Justification: This work does not include theoretical analysis.

    \item {\bf Experimental result reproducibility}
    \item[] Question: Does the paper fully disclose all the information needed to reproduce the main experimental results of the paper to the extent that it affects the main claims and/or conclusions of the paper (regardless of whether the code and data are provided or not)?
    \item[] Answer: \answerYes{} % Replace by \answerYes{}, \answerNo{}, or \answerNA{}.
    \item[] Justification: The "Experiments" section details the experimental setup, including GPU configurations, learning rates, optimizers, and the datasets used. Additionally, the code will be made publicly available upon acceptance.

\item {\bf Open access to data and code}
    \item[] Question: Does the paper provide open access to the data and code, with sufficient instructions to faithfully reproduce the main experimental results, as described in supplemental material?
    \item[] Answer: \answerNo{} % Replace by \answerYes{}, \answerNo{}, or \answerNA{}.
    \item[] Justification: The code is currently proprietary. We intend to open-source it upon acceptance.

\item {\bf Experimental setting/details}
    \item[] Question: Does the paper specify all the training and test details (e.g., data splits, hyperparameters, how they were chosen, type of optimizer, etc.) necessary to understand the results?
    \item[] Answer: \answerYes{} % Replace by \answerYes{}, \answerNo{}, or \answerNA{}.
    \item[] Justification: The settings are specified in the "Experiment" section.

\item {\bf Experiment statistical significance}
    \item[] Question: Does the paper report error bars suitably and correctly defined or other appropriate information about the statistical significance of the experiments?
    \item[] Answer: \answerNo{} % Replace by \answerYes{}, \answerNo{}, or \answerNA{}.
    \item[] Justification: The experiment works with large datasets, and the resources needed are such that running it multiple times for statistical significance would present a significant cost challenge.

\item {\bf Experiments compute resources}
    \item[] Question: For each experiment, does the paper provide sufficient information on the computer resources (type of compute workers, memory, time of execution) needed to reproduce the experiments?
    \item[] Answer: \answerYes{} % Replace by \answerYes{}, \answerNo{}, or \answerNA{}.
    \item[] Justification: We have included the information in the "Experiments" section.
    
\item {\bf Code of ethics}
    \item[] Question: Does the research conducted in the paper conform, in every respect, with the NeurIPS Code of Ethics \url{https://neurips.cc/public/EthicsGuidelines}?
    \item[] Answer: \answerYes{} % Replace by \answerYes{}, \answerNo{}, or \answerNA{}.
    \item[] Justification: We have followed the NeurIPS Code of Ethics.

\item {\bf Broader impacts}
    \item[] Question: Does the paper discuss both potential positive societal impacts and negative societal impacts of the work performed?
    \item[] Answer: \answerYes{} % Replace by \answerYes{}, \answerNo{}, or \answerNA{}.
    \item[] Justification: We have pointed out this work could be used to generate deepfake information.
    
\item {\bf Safeguards}
    \item[] Question: Does the paper describe safeguards that have been put in place for responsible release of data or models that have a high risk for misuse (e.g., pretrained language models, image generators, or scraped datasets)?
    \item[] Answer: \answerNA{} % Replace by \answerYes{}, \answerNo{}, or \answerNA{}.
    \item[] Justification: As we only train our model using standardized NVS datasets, this does not apply to our work. 

\item {\bf Licenses for existing assets}
    \item[] Question: Are the creators or original owners of assets (e.g., code, data, models), used in the paper, properly credited and are the license and terms of use explicitly mentioned and properly respected?
    \item[] Answer: \answerYes{} % Replace by \answerYes{}, \answerNo{}, or \answerNA{}.
    \item[] Justification: We explicitly stated the licenses for both datasets used and included references to their associated research papers. 

\item {\bf New assets}
    \item[] Question: Are new assets introduced in the paper well documented and is the documentation provided alongside the assets?
    \item[] Answer: \answerNA{} % Replace by \answerYes{}, \answerNo{}, or \answerNA{}.
    \item[] Justification: No new datasets or code are being submitted in conjunction with this manuscript.

\item {\bf Crowdsourcing and research with human subjects}
    \item[] Question: For crowdsourcing experiments and research with human subjects, does the paper include the full text of instructions given to participants and screenshots, if applicable, as well as details about compensation (if any)? 
    \item[] Answer: \answerNA{} % Replace by \answerYes{}, \answerNo{}, or \answerNA{}.
    \item[] Justification: This paper does not involve crowdsourcing nor research with human subjects.

\item {\bf Institutional review board (IRB) approvals or equivalent for research with human subjects}
    \item[] Question: Does the paper describe potential risks incurred by study participants, whether such risks were disclosed to the subjects, and whether Institutional Review Board (IRB) approvals (or an equivalent approval/review based on the requirements of your country or institution) were obtained?
    \item[] Answer: \answerNA{} % Replace by \answerYes{}, \answerNo{}, or \answerNA{}.
    \item[] Justification: This paper does not involve crowdsourcing nor research with human subjects.

\item {\bf Declaration of LLM usage}
    \item[] Question: Does the paper describe the usage of LLMs if it is an important, original, or non-standard component of the core methods in this research? Note that if the LLM is used only for writing, editing, or formatting purposes and does not impact the core methodology, scientific rigorousness, or originality of the research, declaration is not required.
    %this research? 
    \item[] Answer: \answerNA{} % Replace by \answerYes{}, \answerNo{}, or \answerNA{}.
    \item[] Justification: This paper does not
involve LLMs as any important, original, or non-standard components

\end{enumerate}

\end{document}